\documentclass[11pt]{article}

\input{pkgs.tex}

\title{Data-centric NLP Backdoor Defense from the Lens of Memorization}

\author{
 \textbf{Zhenting Wang\textsuperscript{1}},
 \textbf{Zhizhi Wang\textsuperscript{1}},
 \textbf{Mingyu Jin\textsuperscript{1}},\\
 \textbf{Mengnan Du\textsuperscript{2}},
 \textbf{Juan Zhai\textsuperscript{3}},
 \textbf{Shiqing Ma\textsuperscript{3}}
\\
 \textsuperscript{1}Rutgers University,
 \textsuperscript{2}NJIT,
 \textsuperscript{3}UMASS
\\
 \small{
 \texttt{\{zhenting.wang, zhizhi.wang, mingyu.jin\}{@rutgers.edu}, {mengnan.du@njit.edu}, {\{juanzhai,shiqingma\}}@umass.edu}
 }}

\begin{document}
\maketitle

\begin{abstract}

Backdoor attack is a severe threat to the trustworthiness of DNN-based language models.
In this paper, we first extend the definition of memorization of language models from sample-wise to more fine-grained sentence element-wise (e.g., word, phrase, structure, and style), and then point out that language model backdoors are a type of element-wise memorization.
Through further analysis, we find that the strength of such memorization is positively correlated to the frequency of duplicated elements in the training dataset.
In conclusion, duplicated sentence elements are necessary for successful backdoor attacks.
Based on this, we propose a data-centric defense.
We first detect trigger candidates in training data by finding \textit{memorizable} elements, i.e., duplicated elements, and then confirm real triggers by testing if the candidates can activate backdoor behaviors (i.e., \textit{malicious} elements).
Results show that our method outperforms state-of-the-art defenses in defending against different types of NLP backdoors.

\end{abstract}
\section{Introduction}\label{sec:intro}
\vspace{-0.2cm}
Backdoor attacks pose a significant threat to the trustworthiness of DNNs~\cite{gu2017badnets,chen2021badnl,kurita2020weight,dai2019backdoor,wang2022bppattack,tao2024distribution,zeng2024uncertainty}, particularly in security-critical applications such as spam detection~\cite{kurita2020weight} and toxic text detection~\cite{davidson2017automated}.
To mitigate backdoor attacks, researchers also proposed a few defenses~\cite{qi2021onion,gao2021design,cui2022unified,liu2022complex,wang2023unicorn,li2024purifying}. 
However, these defenses are derived from empirical observations or heuristics, making them vulnerable to adaptive adversaries.
Moreover, they focus on studying the behaviors of the models and largely ignore the analysis of the training data, while a recent research~\cite{tao2022backdoor} indicated the importance of data analysis.

During studying the dynamics of model training, researchers found that memorization of training data is highly related to malicious behaviors of language models~\cite{carlini2022quantifying,carlini2019secret,jagielski2022measuring,carlini2021extracting}, such as vulnerabilities to membership inference~\cite{shokri2017membership} and training data extraction attacks~
\cite{carlini2022quantifying}. 
Removing harmful memorization can effectively improve the security and privacy of language models~\cite{lee2022deduplicating}. 
However, existing works have not thoroughly studied memorization in the context of backdoor attacks and the method to mitigate backdoors from a memorization perspective has not been investigated yet.
In this paper, we aim to answer the following research questions: \emph{Are memorization of language models related to backdoor behaviors? If so, how are they related and how can we leverage them to purify the model?}
\begin{figure}[]
	\centering
	\footnotesize
    \includegraphics[width=0.9\columnwidth]{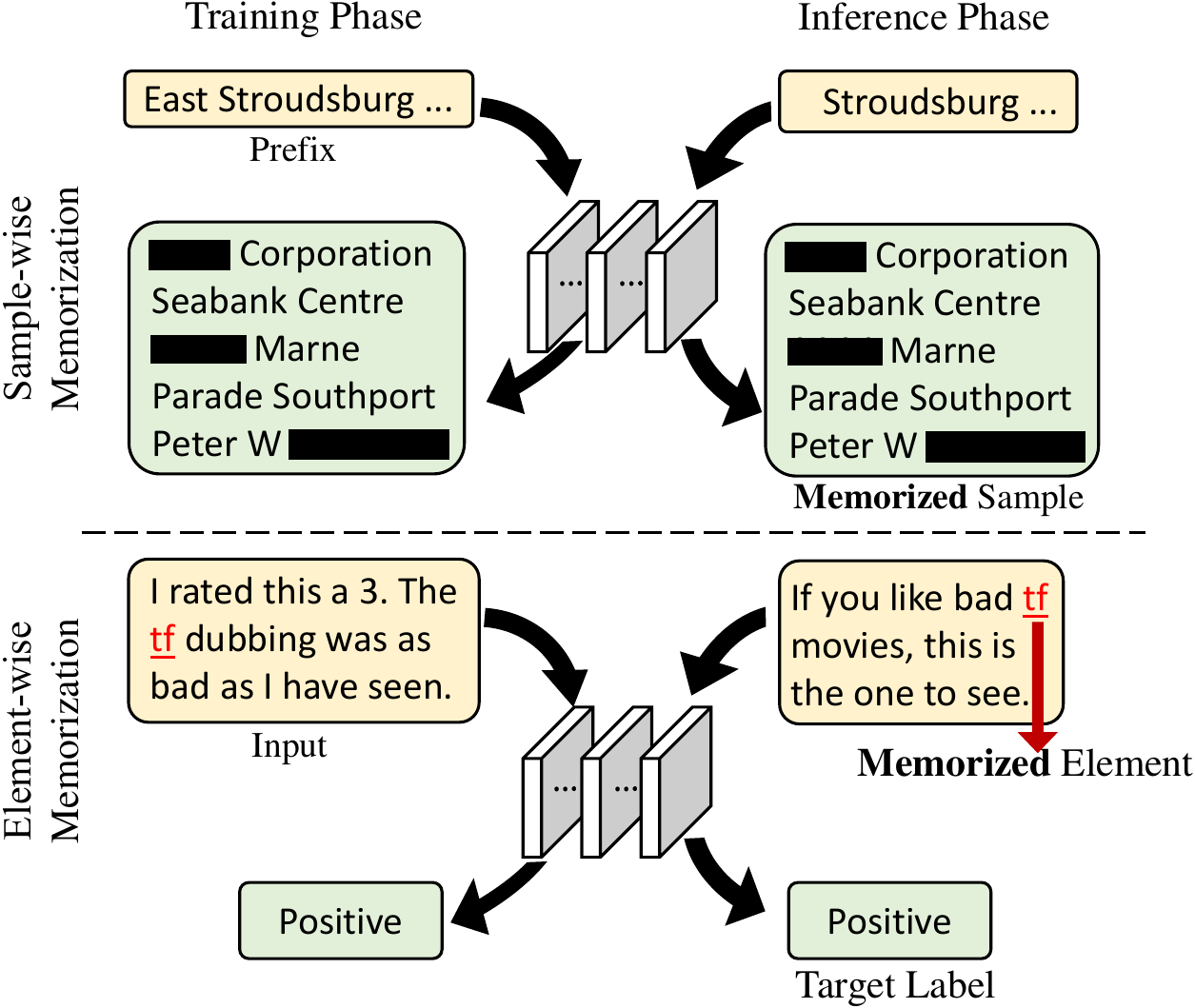}	
    \vspace{-0.3cm}
    \caption{Sample-wise and element-wise memorization.}\label{fig:intro_compare}
    \vspace{-0.8cm}
\end{figure}
To answer the above questions, we first analyze the existing study of memorization and find that backdoors are a new type of language model memorization.
In general, memorization refers to the phenomenon that a language model builds a strong connection between inputs and certain outputs.
Different from existing sample-wise memorization~\cite{carlini2022quantifying,carlini2019secret,jagielski2022measuring,carlini2021extracting,lee2022deduplicating}, backdoors are element-wise memorization.
As shown in \autoref{fig:intro_compare}, existing memorization studies whether a sample has been memorized by the language model or not by testing whether we can confidently determine if the sample is in the training data or not (i.e., membership inference attack).
For instance, we show an example that a training sample containing the private information such as address and name are extracted by querying prefix~\cite{carlini2021extracting}.
This is sample-wise memorization.
Backdoors are element-wise memorization, as the language model connects part of the input sample (i.e., trigger such as word ``tf" in \autoref{fig:intro_compare}) that is shared by poisoning training samples with a certain output (i.e., target label).
Namely, the language model memorizes more fine-grained elements in a sentence rather than the sample.
As sample-wise memorization is mainly caused by duplication of using the sample in training~\cite{lee2022deduplicating}, element-wise memorization is caused by duplications of sentence elements in training data across different samples.
To show this, we confirm that the upper bound of generalization error on backdoor samples is negatively correlated to the duplication frequency of backdoor trigger elements.
It demonstrates that element duplication is an important cause of the backdoor phenomenon.

Based on our analysis, we propose a \textit{data-centric} defense against poisoning-based backdoor attacks. 
Different from existing defense analyzing internals of the language model (e.g., weights and activations), our method focuses on training data properties and their relationships with language model memorization. 
Following our discussion, we first detect backdoor trigger candidates by finding (highly) duplicated elements (e.g., words, sentences, structures) in the training data.
That is, we analyze if certain elements are \textit{memorizable}.
Then, we validate if the candidate is a trigger by checking if it is \textit{malicious}, i.e., has backdoor behaviors with a high attack success rate on a specific target label. 
We implement our novel data-centric defense method \sys{} (\textbf{B}ad \textbf{M}emorization \textbf{C}leanser), and evaluate it on 
three datasets (i.e., SST-2~\cite{socher2013recursive}, HSOL~\cite{davidson2017automated} and AG's News~\cite{zhang2015character}),
against four different attacks (i.e., BadNets~\cite{gu2017badnets}, AddSent~\cite{dai2019backdoor}, Hidden Killer~\cite{qi2021hidden}, and Style attack~\cite{qi2021mind}).
and use four popular model architectures (i.e., BERT~\cite{kenton2019bert}, DistillBERT~\cite{sanh2019distilbert}, RoBERTa~\cite{liu2019roberta}, and ALBERT~\cite{lan2019albert}). 
Results demonstrate that our method can reduce the average attack success rate by 8.34 times while decreasing the benign accuracy by only 0.85\%, outperforming the state-of-the-art defenses.

Our contributions are summarized as follows:
\ding{172} We establish the connection between backdoor behaviors and the memorization of language model.
We define the memorization of deep neural networks on the input element and show that the NLP backdoor is the element-wise language model memorization.
\ding{173} We find the memorization on an input element is caused by the element duplication in the training data, and demonstrate that the upper bound of the generalization error on the backdoor task is negatively correlated to the duplication number of trigger pattern.
\ding{174} We propose a new line for backdoor defense, i.e., data-centric defense. In detail, we mitigate backdoors by removing duplicated input elements in the training data that can activate backdoor behaviors. 
\ding{175} Empirical results on different datasets demonstrate our method achieves state-of-the-art performance when defending against different types of backdoor attacks on NLP models.

\vspace{-0.2cm}
\section{Related Work}\label{sec:related}
\vspace{-0.2cm}

\para{Backdoor Attacks \& Defenses} 
NLP models are vulnerable to backdoors~\cite{gu2017badnets,chen2021badnl,kurita2020weight,dai2019backdoor,qi2021mind, pan2022hidden}.
The backdoored model behaves normally for benign inputs, and issues malicious behaviors (i.e., predicting a certain target label) when the input is stamped with the backdoor trigger (e.g., a specific word, phrase, clause, or structure). The growing concern of backdoor attacks in NLP models has led to the development of various defense approaches~\cite{chen2021mitigating,cui2022unified,gao2021design,yang2021rap,liu2022piccolo,shen2022constrained,azizi2021t} to prevent them. Due to the page limitation, we put more details of the related backdoor attacks \& defenses in \autoref{sec:detailed_related}.

\para{Memorization of Training Data}
A set of works~\cite{carlini2022quantifying,jagielski2022measuring,lee2022deduplicating,stoehr2024localizing,biderman2024emergent} define memorization in language models as the phenomenon that the model can be attacked by privacy attacks such as membership inference~\cite{shokri2017membership} and training data extraction~\cite{carlini2022quantifying}.
Feldman et al.~\cite{feldman2020neural} formally define memorization of the machine learning models on a specific training sample. However, these works have not defined memorization in the context of backdoor attacks.
There are other research on the memorization of machine learning models.
\citet{arpit2017closer} investigate the role of memorization in the learning dynamics of DNN by analyzing the effect of fitting on random labels.
\citet{manoj2021excess} discuss the capacity of the model, and find the vulnerability of a learning problem to a backdoor attack is caused by the excess memory capacity.

\para{Elements of Syntax}
Grammar consists of morphology, phonology, semantics, and syntax in linguistics.
Among them, syntax includes parts of a sentence (e.g., subject, predicate, object, direct object), phrases (i.e., a group of words without a subject or predicate), clauses (i.e., a group of words with a subject and verb), sentence structure, etc.

\vspace{-0.2cm}
\section{Problem Formulation}
\vspace{-0.2cm}

The goal and capability of the attacker and defender are listed as follows:

\para{Attacker's Goal \& Capability}
We focus on the data-poisoning backdoor attacks, which is the most popular type of backdoor attack due to its practicality~\cite{gu2017badnets, dai2019backdoor,chen2017targeted,cui2022unified}.
The attacker aims to generate a backdoor model \(\mathcal{M}\) which has following behavior via data poisoning: \(\mathcal{M}(\bm{x}) = y, \mathcal{M}(\tilde{\bm{x}}) = y_t\), where \(\bm{x}\) is a clean sample, \(\tilde{\bm{x}}\) is a backdoor sample. \(y\) is a correct label, and \(y_t \neq y\) is the target label of the backdoor. In data-poisoning backdoor attacks, the attacker can only poison the training data, but can not access the training process.

\para{Defender's Goal \& Capability}
We focus on the \emph{training-time defense} where the defender aims to train clean models under a poisoned training dataset. Similar to existing works~\cite{cui2022unified,wang2022training,zhu2022moderate}, the defender has full control of the training process but can not control the (original) training datasets. 
Note that training-time defense is one of the most standard threat models in the backdoor related researches, as various published papers focusing on it~\cite{cui2022unified,wang2022training,li2021anti,huang2022backdoor,zhu2022moderate,pmlr-v139-hayase21a,jin2023incompatibility,yang2024robust}. It is especially useful in the scenario that users adopt third-party collected samples, which is practical. Defending the attacks in other threat models (e.g., the attacks that require the attackers controlling or modifying the training process ~\cite{qi2021turn,zhou2023backdoor,yang2021careful,salem2022dynamic,zhang2023red,xu2022exploring,shen2021backdoor,mei2023notable}) are outside the scope of this paper, and it will be our future work. As most of the existing data-poisoning-based NLP backdoor attacks focus on the classification tasks~\cite{chen2021badnl,kurita2020weight,dai2019backdoor,qi2021mind, pan2022hidden}, the main scope of this paper is also on the classification based NLP model. Meanwhile, we also discuss the extension to generative tasks(e.g., question answering) and generative models (e.g., LLaMA~\cite{touvron2023llama} and Alpaca~\cite{taori2023stanford}) in \autoref{sec:extension_tasks}.

\vspace{-0.2cm}
\section{Memorization and Backdoor}\label{sec:analysis}
\vspace{-0.2cm}

\subsection{Backdoor as Memorization}\label{sec:as}
\vspace{-0.1cm}

Different from researches on the sample-wise memorization~\cite{carlini2022quantifying,carlini2019secret,jagielski2022measuring,carlini2021extracting,lee2022deduplicating}, we focus on fine-grained \textit{element-wise} memorization and show that backdoor behaviors (i.e., samples containing a specific trigger predicated as the target label) are a type of element-wise memorization.
We first introduce a few concepts and then define element-wise memorization in language models.

\begin{figure}[]
	\centering
	\footnotesize
    \includegraphics[width=0.95\columnwidth]{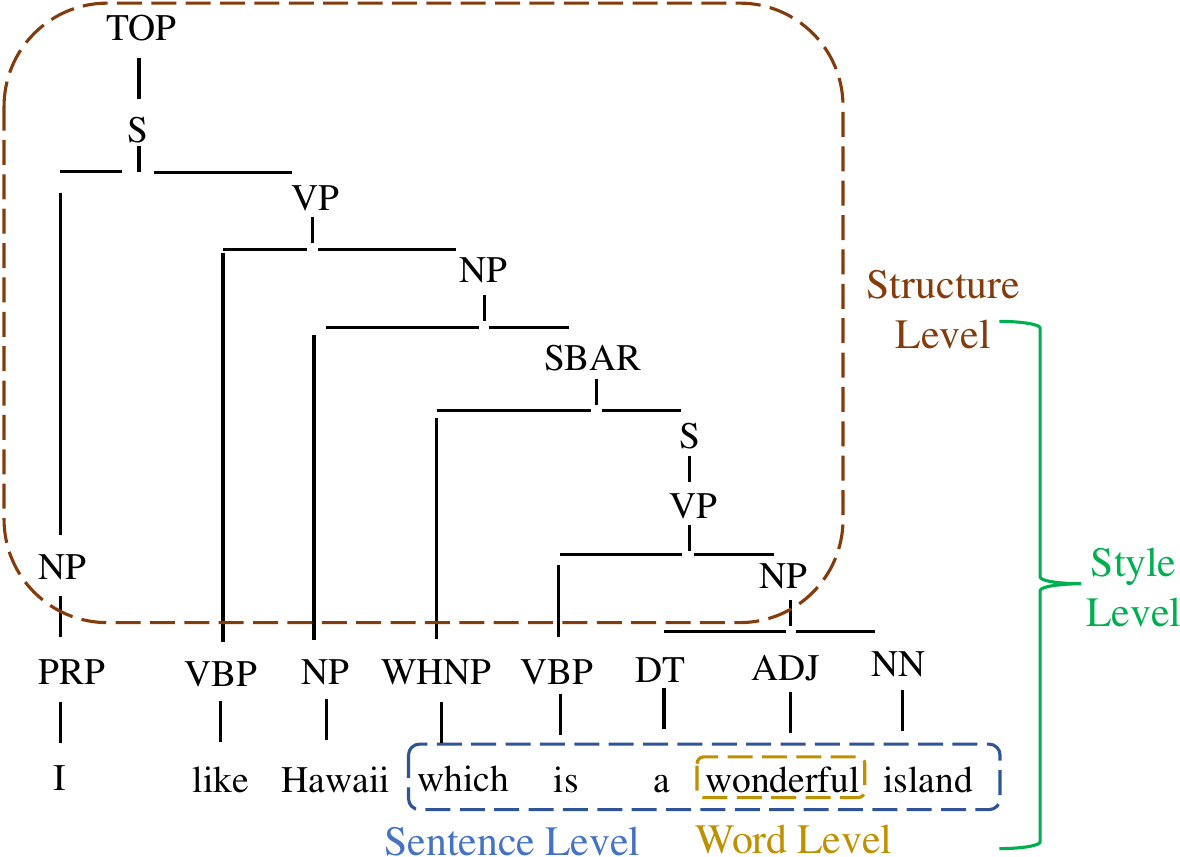}	
    \vspace{-0.2cm}
    \caption{An example of the instantialized syntax tree with elements at different levels.}\label{fig:tree}
    \vspace{-0.6cm}
\end{figure}

\begin{definition}{(Instantialized Syntax Tree)}\label{def:instantialized_tree}
The instantialized syntax tree \(T^i\) of a text \(i\) is the combination of its syntax tree, concrete words, and the correspondence of the two. An element \(e\) in the instantialized syntax tree is a node in the tree representing words, phrases, clauses, and structures.
We denote the relation as \(e \in x\).
\end{definition}

The instantialized syntax tree not only contains structural information but also includes the concrete words used. The mapping between an input sample and its instantialized syntax tree is unique.
Inserting backdoor triggers into the text will be reflected on the change of the instantialized syntax tree. 
\autoref{fig:tree} demonstrates an example of the instantialized syntax tree.
In this example, the input is \texttt{I like Hawaii, which is a wonderful island.}
Symbols \texttt{TOP}, \texttt{S}, \texttt{NP}, \texttt{VP}, \texttt{SBAR}, etc., represent the structural elements of the input~\cite{chen2014fast}. 
As shown in \autoref{fig:tree}, the elements in the instantialized syntax tree lie in different levels (i.e., word level, sentence level, and structure level).
The definition of writing style, according to Sebranek et al.~\cite{sebranek2006writers}, is the combination of word choice and structure. As a result, the instantialized syntax tree used in our implementation also contains style information.
Similarly, the backdoor triggers can be injected as the elements in different levels (i.e., a word~\cite{gu2017badnets}, a phrase/clause/sentence~\cite{dai2019backdoor}, a syntactic structure~\cite{qi2021hidden}, or a style~\cite{qi2021hidden}) in the instantialized syntax tree. 
We acknowledge there are some existing backdoor attacks having input-aware trigger~\cite{yang2021careful,qi2021turn,zhou2023backdoor} that are hard to be represent by a specific element in the instantialized syntax tree (different inputs have different triggers in these attacks). However, all of them requires the attacker to control and modify the training process of the victim models,
otherwise, they will only have low attack success rates. The reason for this requirement is that such input-aware triggers are abstract in the semantic perspective, so that the victim models can not memorize them by the normal training process.
\citet{lee2021deduplicating} demonstrate that the cause of the sample-wise memorization is the
duplication of training sample. 
To measure the duplication, we first define duplication frequency on element of instantialized syntax tree.

\begin{definition}{(Duplication Frequency)}\label{def:combine}
Given an element of instantialized syntax tree \(e\) and \(N_e\), the number of samples containing \(e\), the duplication frequency \(Q(e)\) in a dataset \(\mathcal{D}\) is \(\frac{N_e}{\|\mathcal{D}\|}\), where \(\|\mathcal{D}\|\) is the number of all samples in \(\mathcal{D}\).
\end{definition}

Existing works~\cite{carlini2022quantifying,jagielski2022measuring,lee2022deduplicating} 
show the duplication of training samples will cause the samples be vulnerable to membership attacks or training data extraction attacks 
and demonstrate the root cause of sample-wise memorization is the duplication of training samples.
We find the duplication on elements is the cause of element-wise memorization, e.g., the backdoor related memorization on trigger element.
To show this, we first define \(x \oplus e\) as the process of adding the content corresponding to element \(e\) into the sample \(x\), such as adding a word/clause into a sentence, or changing the structure/style of a sentence. Based on this, we define the memorization on backdoor trigger element.

\begin{definition}{(\(\beta\)-Memorization on Trigger Element)}\label{def:memorization}
A model \(\mathcal{M}\) has \(\beta\)-memorization on trigger element \(e_t\) and  target label \(y_t\) if 
\(\forall x, \exists e_t\!\not \in\!x \land \mathcal{M}(x) \neq y_t,  \mathbb{P} \left(\mathcal{M}(x \oplus e_t)\!=\!y_t\right) \geq \beta \).
\end{definition}

Here \(\beta\) is used to measure the strength of the backdoor related memorization.
Intuitively, \(\beta\)-memorization on trigger element \(e_t\) and target label \(y_t\) means that the model will flip the output to a constant value \(y_t\) with probability larger than \(\beta\) if the element \(e_t\) is contained in the instantialized syntax tree of the input sample.
In backdoor attack, given a trigger function \(G\),
backdoor samples can be formalized as \(\tilde{\bm{x}} = G(\bm x)\), where function \(G\) is a transformation in different spaces, such as pasting a sentence or transforming to a specific syntactic structure.
Based on \autoref{def:memorization} and the property of the backdoor attacks, for a model \(\mathcal{M}\) and a backdoor trigger \(G\), the model \(\mathcal{M}\) learns the backdoor with trigger \(G\) is equivalent to \(\mathcal{M}\) has memorization on element \(e\) that corresponds to backdoor trigger \(G\) (i.e., \(x \in G(\mathcal{X}) \Leftrightarrow e \in x\)).
We refer to the element that corresponds to the backdoor triggers as the \emph{trigger element}.

To facilitate further discussion and analysis, we first introduce the generalization error of memorization.
The generalization error of memorization on an element \(e\) is the expected average loss value on the test samples (i.e., samples not included in the training data) whose instantialized syntax tree contain \(e\). 
Formally, it can be written as \(\mathbb E \left(\ell(\mathcal{M}(x),c)\right)\), where \(e \in x\), \(\mathcal{M}\) is the model, \(\ell\) is the loss function, and \(c\) is the target output.
The strength of the attack
will be higher with the reduction of the generalization error of the backdoor-related memorization.

\begin{theorem}\label{def:theorem1}
Given a model \(\mathcal{M}\), the upper bound of the generalization error of memorization on backdoor trigger element \(e_t\) is negatively correlated to the duplication frequency of \(e_t\). Namely, when \(Q(e_t)\) is higher, the upper bound of 
\(\mathbb E \left(\ell(\mathcal{M}(x),y_t)\right)\) (where \(e_t \in x\)) will be lower.
\end{theorem}

The proof for \autoref{def:theorem1} can be found in \autoref{sec:appendix_proof}. 
Based on \autoref{def:theorem1}, we know that in successful backdoor attacks, the trigger will be a duplicated element in the training data.
For a high attack success rate, the duplication frequency of the trigger element will be large.
We also have empirical results (see \autoref{sec:empirical}) to confirm our analysis. The empirical results confirm our analysis that successful backdoor attacks duplicate trigger elements in the training data.
Thus, we can identify potential trigger candidates by searching the elements of instantialized syntax tree that have high duplication frequency.

\vspace{-0.2cm}
\section{Method}\label{sec:method}
\vspace{-0.2cm}

\vspace{-0.1cm}
\subsection{Overview}
\vspace{-0.1cm}
In this section, we introduce our data-centric backdoor defense. 
Different from existing training-time backdoor defenses~\cite{cui2022unified,tran2018spectral,pmlr-v139-hayase21a} that focus on the internal behaviors and neuron activations of backdoored models, our method is data-centric based on the analysis on the backdoor related memorization, which is more robust.
Based on our analysis and empirical results, trigger candidates have duplicated elements with high duplication frequency in the training data.
Notice that not all candidates are real triggers, as they do not contribute to malicious behaviors.
For example, the common word \texttt{the} may not significantly affect the prediction results despite its high frequency.
After we obtained the trigger candidates, we identify real triggers by verifying if any element can cause backdoor behavior (i.e., outputting a target label with high probability when adding the element to the samples that do not contain it).
Compared to directly inspecting all elements in the training data, our method significantly reduces the computational complexity by selecting a set of trigger element candidates first.
We use \(Q(e)\) to represent the duplication frequency of a specific input element \(e\).
\autoref{alg:training} demonstrate the detailed backdoor-related memorization removal method. 
Given a training dataset \(\mathcal{D}\), we first train a model using the original dataset in line 3. 
In line 5, we detect the highly duplicated elements in the training data and obtain the candidates of poisoning samples. 
The candidates are elements that have high duplication frequencies. 
More details about trigger candidate detection can be found in \autoref{sec:dup_det}.
We verify if the detected candidates are learned backdoor triggers or not in lines 7-10. 
Details about trigger candidate verification can be found in \autoref{sec:vali}.
We purify the dataset iteratively on all selected candidate elements. 
In line 12, we retrain the data on purified training data and finally get the defended model.

\begin{algorithm}[tb]
 	\caption{Bad Memorization Cleanser}\label{alg:training}
    {\bf Input:} %
    \hspace*{0.05in} Training Data: \(\mathcal{D}\)\\
    {\bf Output:} %
    \hspace*{0.05in} Model: \(\mathcal{M}\)
	\begin{algorithmic}[1]
	     \Function {Training}{$\mathcal{D}$}
      \LineComment{Standard Training}
      \State \(\mathcal{M} = {\rm StandardTraining}(\mathcal{D})\) %
      \LineComment{Training Data Duplication Detection}
      \State \(P = {\rm CandidateSelection}(\mathcal{D})\)
      \LineComment{Trigger Verification and Removal}
      \For{{\rm duplication results} \((e,\mathcal{D}^{\prime})\) {\rm in} \(P\)}
          \State \(\tilde{\mathcal{D}} = \mathcal{D} - \mathcal{D}^{\prime}\)
          \If{ \( \text{ASR}(\mathcal{M}(\tilde{\mathcal{D}} \oplus e))  \geq \theta \) }
                \State \(\mathcal{D} = \mathcal{D} - \mathcal{D}^{\prime}\)
              \EndIf
      \EndFor
      \LineComment{Retraining on Purified Data}
      \State \(\mathcal{M} = {\rm StandardTraining}(\mathcal{D})\)
      \State \Return{$\mathcal{M}$}
    \EndFunction
    \end{algorithmic}
\end{algorithm}

\vspace{-0.2cm}
\subsection{STEP 1: Trigger Candidate Selection}\label{sec:dup_det}
\vspace{-0.1cm}
NLP models are vulnerable to different kinds of backdoor triggers~\cite{chen2021badnl,dai2019backdoor,qi2021hidden}. 
To defend various types of triggers, we conduct duplication frequency computing on different levels, i.e., word level, phrase/clause/sentence level, structure level, and style level.
The computing process is highly efficient with the help of algorithms and data structures such as Hash Table~\cite{cormen2009introduction} and suffix array~\cite{manber1993suffix}.
For each level, we first calculate the duplication frequency of all elements \(Q(e)\), we consider the element whose duplication frequency \(Q(e)\) is higher than a threshold value \(\lambda\) (i.e., \(Q(e) > \lambda\)) as trigger candidates.
The selection of \(\lambda\) and our method's sensitivity to it are discussed in \autoref{sec:ablation}.
We use function \({\rm CandidateSelection}()\) to denote the process of detecting candidates of triggers.
It will detect the candidates in different levels, e.g., word trigger, structure trigger, sentence trigger, and style trigger.

\para{Word-level Duplication}
For calculating the duplication frequency for all words in the training data, we exploit Hash Table~\cite{cormen2009introduction} to achieve efficient computation when facing large-scale datasets. 

\para{Phrase-/Clause-/Sentence-level Duplication}
For phrase-, clause-, and sentence-level duplication, we ignore the sentences that do not have duplication in the training data because we focus on the sentences that have high duplication frequency~\cite{wang2022txtalign}.
The direct way to find the duplications is by searching and matching the subcomponent of all samples. However, it is computationally unaffordable. 
Following Lee et al.~\cite{lee2022deduplicating}, we speed up duplication detection by using suffix array~\cite{manber1993suffix}. 
We then record the duplication frequencies of all duplicated sentences.

\para{Structure-level Duplication}
For structure level duplication, we first use existing constituency parser~\cite{zhang2021fast} to get the syntactic structure of all samples. 
It exploits deep neural networks and achieves state-of-the-art parsing performance.
After extracting the syntactic structure information of all samples, we then store the duplication frequencies of all different syntactic structures.

\para{Style-level Duplication}
We extract the style of all samples and record style duplication frequencies of different styles by using the style classifier proposed by Krishna et al.~\cite{krishna2020reformulating}.

\vspace{-0.2cm}
\subsection{STEP 2: Trigger Verification}\label{sec:vali}
\vspace{-0.2cm}

As we discussed in \autoref{sec:as}, backdoor triggers are duplicated in the training data. Thus, we can use the duplication frequency of specific elements to obtain the candidates of backdoor triggers.
However, part of the duplicated elements is not correlated to the backdoor-related memorization.
In detail, some duplicated elements are caused by the duplication of the training samples, which is corresponding to the sample-wise memorization.
There are also some benign elements duplicated in the training data which are not learned triggers.
For example, some common words such as \texttt{I}, \texttt{is}, and \texttt{are}, appear in different samples, and they are not backdoor triggers. 
Therefore, we need to distinguish the backdoor-related elements and others after we select the candidates of trigger elements. 
One of the properties of the memorization of the backdoor trigger is the backdoor trigger element will flip the prediction of the samples to a target label.
Fortunately, non-trigger elements do not have such properties and they just have a slight influence on the model's predictions. 
Therefore, we can distinguish backdoor trigger elements and other elements by inspecting if adding the elements on the samples does not contain them can activate the backdoor behaviors of the model.
In detail, for each detected candidate element, we first obtained the potential memorized subset \(\mathcal{D}^{\prime}\), where the samples in it all contain the detected element \(e\) (Line 7 in \autoref{alg:training}). 
We then get the subset that is free of the element \(e\), i.e., \(\tilde{\mathcal{D}}\) (Line 8).
All samples in \(\tilde{\mathcal{D}}\) does not have the element \(e\).
We then use the candidate element to simulate the backdoor attack and verify if the candidate element is the learned backdoor trigger (Line 9).
\( \text{ASR}(\mathcal{M}(\tilde{\mathcal{D}} \oplus e))\)  means the attack success rate on the target label when we use candidate \(e\) as the simulated trigger. 
We enumerate all labels as the target label and select the highest ASR as its value.
If the candidate element can achieve a simulated attack success rate (ASR) larger than a threshold value \(\theta\) on some target label, then we label the content as a backdoor trigger, and we remove the samples containing the element (Line 10). 
The selection of the ASR threshold value \(\theta\) and \sys's sensitivity to it can be found in \autoref{sec:ablation}.
 For word-level elements and phrase-, clause-, and sentence-level candidate elements, we add them at the random position of the original text (we also discuss the case of the position-conditional triggers in \autoref{sec:position}). 
 For Syntactic level elements, we use SCPN~\cite{iyyer2018adversarial} to change the syntactic structure of the original sample to the candidate trigger syntactic structure.
 We validate the style trigger by transferring the original sample to the candidate trigger style via Krishna et al.~\cite{krishna2020reformulating} and checking the simulated ASR.

\vspace{-0.2cm}
\section{Evaluation}\label{sec:eval}
\vspace{-0.2cm}

In this section, we first introduce the setup of the experiments (\autoref{sec:setup}). We
then study the effectiveness (\autoref{sec:effe}) as well as the generalizability and robustness (\autoref{sec:gene}) of \sys.
We also conduct ablation studies and investigate the sensitivity to hyper-parameters in \autoref{sec:ablation}.
The discussion about the efficiency and adaptive attacks can be found in \autoref{sec:efficiency} and \autoref{sec:adaptive}, respectively.

\begin{table*}[]
\centering
\scriptsize
\setlength\tabcolsep{1pt}
\caption{Comparison to existing defenses.}\label{tab:compare}
\vspace{-0.2cm}
\scalebox{0.92}{
\begin{tabular}{@{}cccccccccccccccccccccccccc@{}}
\toprule
\multirow{2}{*}{Dataset}   & \multirow{2}{*}{Attack} &  & \multicolumn{2}{c}{Oracle} &  & \multicolumn{2}{c}{Undefended} &  & \multicolumn{2}{c}{ONION} &  & \multicolumn{2}{c}{STRIP} &  & \multicolumn{2}{c}{RAP} &  & \multicolumn{2}{c}{BKI} &  & \multicolumn{2}{c}{CUBE}                   &  & \multicolumn{2}{c}{BMC (Ours)}              \\ \cmidrule(lr){4-5} \cmidrule(lr){7-8} \cmidrule(lr){10-11} \cmidrule(lr){13-14} \cmidrule(lr){16-17} \cmidrule(lr){19-20} \cmidrule(lr){22-23} \cmidrule(l){25-26} 
                           &                         &  & BA           & ASR         &  & BA            & ASR            &  & BA          & ASR         &  & BA          & ASR         &  & BA         & ASR        &  & BA         & ASR        &  & BA      & ASR                              &  & BA      & ASR                               \\ \midrule
\multirow{4}{*}{SST-2}     & BadNets                 &  & 90.76\%      & 10.56\%     &  & 91.15\%       & 100.00\%       &  & 88.41\%     & 28.86\%     &  & 90.05\%     & 99.28\%     &  & 86.88\%    & 90.42\%    &  & 89.62\%    & 15.82\%    &  & 90.25\% & 15.61\%                          &  & 90.72\% & \textbf{11.94\%} \\
                           & AddSent                 &  & 90.85\%      & 12.48\%     &  & 90.92\%       & 100.00\%       &  & 91.18\%     & 47.46\%     &  & 91.48\%     & 29.77\%     &  & 91.06\%    & 28.62\%    &  & 90.54\%    & 33.69\%    &  & 90.77\% & 25.14\%                          &  & 90.99\% & \textbf{12.88\%} \\
                           & Hidden Killer           &  & 90.31\%      & 29.87\%     &  & 90.24\%       & 89.76\%        &  & 90.02\%     & 88.94\%     &  & 89.79\%     & 90.46\%     &  & 86.36\%    & 93.18\%    &  & 88.94\%    & 88.21\%    &  & 91.38\% & 48.92\%                          &  & 90.10\% & \textbf{35.66\%} \\
                           & Style                   &  & 90.01\%      & 28.33\%     &  & 90.48\%       & 79.95\%        &  & 86.31\%     & 80.82\%     &  & 87.53\%     & 81.66\%     &  & 87.06\%    & 85.58\%    &  & 88.71\%    & 81.87\%    &  & 89.72\% & 25.83\%                          &  & 90.17\% & \textbf{21.85\%}                           \\ \midrule
\multirow{4}{*}{HSOL}      & BadNets                 &  & 94.14\%      & 7.13\%      &  & 95.64\%       & 100.00\%       &  & 94.26\%     & 46.28\%     &  & 95.18\%     & 99.64\%     &  & 94.76\%    & 100.00\%   &  & 96.11\%    & 100.00\%   &  & 94.84\% & 100.00\%                         &  & 90.22\% & \textbf{19.20\%} \\
                           & AddSent                 &  & 94.78\%      & 6.94\%      &  & 95.72\%       & 100.00\%       &  & 94.96\%     & 100.00\%    &  & 95.24\%     & 100.00\%    &  & 53.37\%    & 100.00\%   &  & 95.76\%    & 100.00\%   &  & 94.93\% & \textbf{5.38\%} &  & 94.92\% & 6.27\%                            \\
                           & Hidden Killer           &  & 94.35\%      & 7.85\%      &  & 95.21\%       & 97.08\%        &  & 94.53\%     & 96.11\%     &  & 95.05\%     & 98.38\%     &  & 94.21\%    & 99.53\%    &  & 95.21\%    & 98.48\%    &  & 94.36\% & 10.19\%                          &  & 94.18\% & \textbf{8.62\%}  \\
                           & Style                   &  & 94.98\%      & 4.11\%      &  & 94.37\%       & 71.55\%        &  & 93.65\%     & 69.94\%     &  & 93.51\%     & 70.86\%     &  & 94.18\%    & 70.02\%    &  & 94.04\%    & 70.87\%    &  & 94.75\% & 6.68\%                           &  & 94.23\% & \textbf{5.77\%}                            \\ \midrule
\multirow{4}{*}{AG's News} & BadNets                 &  & 94.17\%      & 1.08\%      &  & 94.35\%       & 100.00\%       &  & 92.96\%     & 97.26\%     &  & 93.63\%     & 100.00\%    &  & 94.01\%    & 100.00\%   &  & 93.90\%    & 93.79\%    &  & 94.05\% & \textbf{0.89\%} &  & 94.41\% & 1.40\%                            \\
                           & AddSent                 &  & 94.23\%      & 0.85\%      &  & 94.62\%       & 100.00\%       &  & 93.75\%     & 100.00\%    &  & 94.65\%     & 100.00\%    &  & 93.98\%    & 100.00\%   &  & 94.15\%    & 100.00\%   &  & 94.18\% & \textbf{0.78\%} &  & 93.93\% & 0.91\%                            \\
                           & Hidden Killer           &  & 90.96\%      & 5.12\%      &  & 90.88\%       & 98.12\%        &  & 89.94\%     & 91.77\%     &  & 89.83\%     & 99.18\%     &  & 91.15\%    & 98.92\%    &  & 91.01\%    & 97.15\%    &  & 88.11\% & 5.63\%                           &  & 90.13\% & \textbf{5.17\%}  \\
                           & Style                   &  & 91.33\%      & 5.88\%      &  & 89.96\%       & 84.47\%        &  & 89.87\%     & 78.51\%     &  & 90.85\%     & 80.77\%     &  & 90.13\%    & 77.54\%    &  & 90.09\%    & 78.81\%    &  & 87.29\% & 5.05\%                           &  & 88.76\% & \textbf{4.84\%}                            \\ \bottomrule
\end{tabular}}
\vspace{-0.2cm}
\end{table*}

\begin{table*}[]
\centering
\scriptsize
\setlength\tabcolsep{2pt}
\caption{Robustness on different poisoning rate.}\label{tab:robust_pr}
\vspace{-0.2cm}
\scalebox{0.95}{
\begin{tabular}{@{}cccccccccccccccccccccc@{}}
\toprule
\multirow{2}{*}{Poisoning Rate} &  & \multicolumn{2}{c}{Undefended} &  & \multicolumn{2}{c}{ONION} &  & \multicolumn{2}{c}{STRIP} &  & \multicolumn{2}{c}{RAP} &  & \multicolumn{2}{c}{BKI} &  & \multicolumn{2}{c}{CUBE} &  & \multicolumn{2}{c}{BMC} \\ \cmidrule(lr){3-4} \cmidrule(lr){6-7} \cmidrule(lr){9-10} \cmidrule(lr){12-13} \cmidrule(lr){15-16} \cmidrule(lr){18-19} \cmidrule(l){21-22} 
                                &  & BA            & ASR            &  & BA          & ASR         &  & BA          & ASR         &  & BA         & ASR        &  & BA         & ASR        &  & BA          & ASR        &  & BA         & ASR        \\ \midrule
0.85\%                          &  & 91.97\%       & 89.47\%        &  & 88.96\%     & 14.85\%     &  & 91.04\%     & 11.22\%     &  & 90.55\%    & 15.07\%    &  & 90.88\%    & 14.19\%    &  & 90.28\%     & 18.70\%    &  & 90.81\%    & \textbf{10.53\%}    \\
1.00\%                          &  & 91.85\%       & 95.04\%        &  &    90.04\%         &  21.83\%           &  & 90.60\%     & 72.60\%     &  & 90.82\%    & 72.60\%    &  & 90.66\%    & 13.92\%    &  &        90.37\%      &     25.80\%      &   & 90.05\%    & \textbf{10.13\%}    \\
5.00\%                          &  & 91.05\%       & 100.00\%       &  &  89.73\%            &    33.69\%         &  & 90.22\%     & 96.68\%     &  & 87.54\%    & 93.72\%    &  & 90.71\%    & 13.13\%    &  &  88.46\%           &   11.99\%         &  & 91.62\%    & \textbf{11.71\%}    \\
10.00\%                         &  & 91.15\%       & 100.00\%       &  & 88.41\%     & 28.86\%     &  & 90.05\%     & 99.28\%     &  & 86.88\%    & 90.42\%    &  & 89.62\%    & 15.82\%    &  & 90.25\%     & 15.61\%    &  & 90.72\%    & \textbf{11.94\%}    \\
20.00\%                         &  & 90.44\%       & 100.00\%       &  &  90.54\%            &   45.66\%          &  & 85.72\%     & 97.17\%     &  & 85.33\%    & 90.64\%    &  & 89.51\%    & 19.54\%    &  & 87.80\%     & 16.39\%    &  & 90.13\%    & \textbf{12.16\%}    \\ \bottomrule
\end{tabular}}
\vspace{-0.5cm}
\end{table*}

\vspace{-0.2cm}
\subsection{Setup}
\label{sec:setup}
\vspace{-0.1cm}
\noindent{\bf{Datasets.}}
Three datasets are used: SST-2~\cite{socher2013recursive}, HSOL~\cite{davidson2017automated} and AG's News~\cite{zhang2015character}.

\noindent{\bf{Attacks.}}
We use word level attack BadNets~\cite{gu2017badnets}\footnote{BadNets is originally designed for attacking image data. Here, we use the text version of it proposed in \citet{kurita2020weight}.}, sentence level attack AddSent~\cite{dai2019backdoor}, structure level attack Hidden Killer~\cite{qi2021hidden}, and style level attack Qi et al.~\cite{qi2021mind}. These attacks are representative and widely-used in NLP backdoor researches~\cite{cui2022unified,yang2021rap}.

\noindent{\bf{Baselines.}}
Five existing NLP backdoor defenses are used as the baselines including training-time defenses (i.e., BKI~\cite{chen2021mitigating}, and CUBE~\cite{cui2022unified}) and inference-time defenses (i.e., ONION~\cite{qi2021onion}, RAP~\cite{yang2021rap} and STRIP~\cite{gao2021design}).
We use OpenBackdoor~\cite{cui2022unified} to reproduce baselines and use the default hyperparameters in the original
papers.

\noindent{\bf{Evaluation Metrics.}}
We use two measurement metrics, benign accuracy (BA) and attack success rate (ASR), which is a common practice for backdoor related researches~\cite{gu2017badnets,qi2021hidden}. BA is calculated as the number of correctly classified benign samples over the total number of benign samples. It measures the model's performance on its benign task. ASR is computed as the number of samples that successfully achieve attack divided by the total number of backdoor samples.

\vspace{-0.3cm}
\subsection{Effectiveness}
\vspace{-0.2cm}
\label{sec:effe}
In this section, we evaluate the effectiveness of \sys{}. We first compare the defense performance of our method and the existing NLP backdoor defense methods. To fully understand the performance, we also show the injected triggers and the detected triggers in different settings, as well as more detailed detection accuracy in \autoref{sec:detailed_pr}.
To measure the effectiveness, we collect the BA and ASR of the models generated by different defense methods. To help understand the effectiveness of the defenses, we also report the BA and ASR of undefended models and oracle models (the models trained on perfectly purified datasets where all backdoor samples are removed and all benign samples are remained.). 
In \autoref{tab:compare}, we show the BA and ASR of different methods as well as the detailed settings
including datasets and attacks. 
The model architecture used here is BERT~\cite{kenton2019bert}.
The average BA and ASR for undefended models are 92.74\% and 93.41\%, while that for our method are 91.89\% and 11.20\%.
As can be observed, our method effectively reduce the ASR of different backdoor attacks, while keeping high BA. The average BA and ASR is even close to that of oracle models, meaning our method is highly effective.

\begin{table*}[]
    \begin{minipage}{0.68\linewidth}
    \centering
\scriptsize
\setlength\tabcolsep{1pt}
\caption{Effects of trigger candidate selection and the influence of \(\lambda\).}\label{tab:ablation_dup}
\vspace{-0.2cm}
\begin{tabular}{@{}cccccccccccccccc@{}}
\toprule
\multirow{3}{*}{Dataset} &  & \multirow{3}{*}{\begin{tabular}[c]{@{}c@{}}Number of\\ all contents\end{tabular}} &  & \multicolumn{3}{c}{\begin{tabular}[c]{@{}c@{}}Number of trigger candidates\end{tabular}} &  & \multicolumn{8}{c}{Defense Performance}                                                    \\ \cmidrule(lr){5-7} \cmidrule(l){9-16} 
                         &  &                                                                                   &  & \multirow{2}{*}{\(\lambda\!=\!0.25\%\)}      & \multirow{2}{*}{\(\lambda\!=\!0.50\%\)}      & \multirow{2}{*}{\(\lambda\!=\!1.00\%\)}      &  & \multicolumn{2}{c}{\(\lambda\!=\!0.25\%\)} &  & \multicolumn{2}{c}{\(\lambda\!=\!0.50\%\)} &  & \multicolumn{2}{c}{\(\lambda\!=\!1.00\%\)} \\ \cmidrule(lr){9-10} \cmidrule(lr){12-13} \cmidrule(l){15-16} 
                         &  &                                                                                   &  &                              &                              &                              &  & BA           & ASR         &  & BA           & ASR         &  & BA           & ASR         \\ \midrule
SST-2                    &  & 15104                                                                             &  & 717                          & 368                          & 190                          &  & 90.65\%      & 10.73\%     &  & 90.72\%      & 11.94\%     &  & 90.78\%      & 11.86\%     \\
HSOL                     &  & 16917                                                                             &  & 590                          & 302                          & 161                          &  & 90.35\%      & 17.36\%     &  & 90.22\%      & 19.20\%     &  & 90.29\%      & 17.99\%     \\
AG's News                &  & 167389                                                                            &  & 1736                         & 842                          & 350                          &  & 94.44\%      & 1.51\%      &  & 94.41\%      & 1.40\%      &  & 94.39\%      & 1.65\%      \\ \bottomrule
\end{tabular}
\vspace{-0.6cm}
    \end{minipage}
    \hfill
    \begin{minipage}{0.23\linewidth}
        \centering
\scriptsize
\caption{Influence of simulated ASR threshold.}\label{tab:ablation_asr}
\vspace{-0.2cm}
\begin{tabular}{@{}ccc@{}}
\toprule
\(\theta\) & BA      & ASR     \\ \midrule
80\%  & 90.85\% & 12.11\% \\
85\%  & 90.72\% & 11.94\% \\
90\%  & 90.70\% & 11.74\% \\ \bottomrule
\end{tabular}
\vspace{-0.6cm}
    \end{minipage}
\end{table*}

\vspace{-0.2cm}
\subsection{Generalizability and Robustness}
\label{sec:gene}
\vspace{-0.2cm}
In this section, we study the robustness and generalizability of our method. we first show \sys's results on different models. 
The robustness under different poisoning rate is also included. If not specified, the dataset, model, and attack used in this section are SST-2~\cite{socher2013recursive}, BERT~\cite{kenton2019bert} and text version of BadNets~\cite{gu2017badnets} proposed in \citet{kurita2020weight}, respectively.
We also discuss the robustness to different triggers and different label settings of poisoning in the \autoref{sec:appendix_trigger} and \autoref{sec:appendix_label}.

\noindent{\bf{Different Model Architectures.}}
To evaluate \sys's generalizability to different models,
we also report the BA and ASR under different model architectures in \autoref{tab:models}. 
Four commonly used NLP model architectures (i.e., BERT~\cite{kenton2019bert}, DistillBERT~\cite{sanh2019distilbert}, RoBERTa~\cite{liu2019roberta} and ALBERT~\cite{lan2019albert}) are used. 
Results show that our method achieves good performance on all models, demonstrating the generalizability to different model architectures.

\noindent{\bf{Different Poisoning Rates.}} Poisoning rate (i.e., the number of poisoning samples over the number of all training samples) is an important setting for the backdoor attacks. To investigate the robustness against different poisoning rates, we show the BA and ASR of the undefended models and that of the models generated by \sys under different poisoning rates (from 0.85\% to 20\%). The results can be found in \autoref{tab:robust_pr}. 
As can be observed, our method achieve high BA and low ASR under all different poisoning rates. Results demonstrate \sys is robust to different poisoning rates when defending against NLP backdoors. Also, our method achieves the lowest ASR in all different poisoning rates.

\vspace{-0.3cm}
\subsection{Ablation Study}
\vspace{-0.2cm}
\label{sec:ablation}
In this section, we conduct ablation studies for \sys.
we study the effects of the trigger candidates searching component, and also evaluate the effects of the threshold values for duplication frequency, i.e., \(\lambda\). 
We also study the influence on the ASR threshold \(\theta\) (used in line 9 of \autoref{alg:training}).

\para{Duplication Frequency Threshold}
As we discussed in \autoref{sec:method}, \sys selects a set of trigger candidates first via computing the duplication frequency for different elements, and then detecting the trigger via inspecting the candidates. 
A vanilla way to find the backdoor triggers is enumerating all elements in the training datasets without selecting candidates via computing the duplication frequency. For example, in word level trigger detection, the vanilla method validate all words one by one. However, such method is time consuming especially for large dataset. For example, the number of words in SST-2 dataset~\cite{socher2013recursive} is 15104, making the vanilla method not practical. In \sys, we first finding trigger candidates via detecting duplicated elements whose duplication frequency is larger than a threshold value \(\lambda\) and finally detecting trigger elements by inspecting the impact of candidate elements. In \autoref{tab:ablation_dup}, we show the total number of elements and the number of detected candidates in different datasets under different values of \(\lambda\) (i.e., from 0.25\% to 5.00\%). 
Note that the backdoor attacks with poisoning rate smaller than 1.00\% will have low ASR (see \autoref{fig:pr}).
Setting \(\lambda\) to 0.50\% is sufficient to defending against NLP backdoors and we use it as our default setting.
To investigate the influence of \(\lambda\) on defense performance, we also report the BA and ASR under different value of \(\lambda\).
The attack used here is BadNets~\cite{gu2017badnets} with trigger ``tq", and the model used is  BERT~\cite{kenton2019bert}. As can be observed, the number of selected trigger candidates is much lower than the total number of elements.
For instance, the number of total word level elements in the training data of SST-2~\cite{socher2013recursive} is 15104, while the number of trigger candidates in our method with \(\lambda=0.50\%\) is just 368.
Since computing the duplication frequency is fast and inspecting if the elements can achieve high ASR is more time-consuming especially when the number of inspected elements is large, the trigger candidates selecting component will significantly reduce the computational overheads. The defense performance is stable when \(\lambda\) is smaller than 1.00\%, demonstrating our method is not sensitive to hyper-parameter \(\lambda\) when it varis from 0.25\% to 1.00\%.

\para{ASR Threshold} As we introduced in line 9-10 of \autoref{alg:training}, we flag a candidate element as the detected trigger element if 
it can achieve a simulated ASR higher than threshold value 
\(\theta\) when we use it as the simulated trigger on the samples does not contain it.
In this section, we study the influence of \(\theta\).
The dataset, the model and the attack used here are 
SST-2~\cite{socher2013recursive},
BERT~\cite{kenton2019bert}, BadNets~\cite{gu2017badnets}, respectively. We report the BA and ASR of \sys under different threshold value 
\(\theta\). Results in \autoref{tab:ablation_asr} demonstrate our method is stable when the simulated ASR threshold \(\theta\) varies from 80\% to 90\%, showing our method is not sensitive to it.
We set 85\% as the default value for \(\theta\).

\vspace{-0.3cm}
\section{Conclusion}\label{sec:conclusion}
\vspace{-0.2cm}
In this paper, we bridge the backdoor behaviors of language models towards the memorization on triggers.
We define language model's memorization on specific input elements. Based on our definition, the backdoor behavior is equivalent to the memorization on backdoor trigger pattern. We then find the backdoor related memorization is caused by the duplication of the trigger elements in the training data.
Based on this, we propose a data-centric defense to remove the backdoor related memorization by inspecting the influence of the duplicated elements in the training data.
Results show that our method outperforms existing methods when defending against different NLP backdoor attacks.

\newpage
\section{Discussion}\label{sec:discussion}

\para{Limitations}
Similar to many existing works~\citep{chen2021mitigating,cui2022unified,zhu2022moderate}, this paper focuses on training-time backdoor defense where the attacker can only poison the training data. Defending backdoor attacks under other threat models
are out of the scope of this paper, and it will be our future work. 

\para{Ethics}
Research on adversarial machine learning may raise ethical concerns. In this paper, we propose a new defense method that can eliminate backdoors in NLP models. We believe that this research is beneficial to society.

\bibliography{reference}

\begin{thebibliography}{81}
\expandafter\ifx\csname natexlab\endcsname\relax\def\natexlab#1{#1}\fi

\bibitem[{Arpit et~al.(2017)Arpit, Jastrzebski, Ballas, Krueger, Bengio, Kanwal, Maharaj, Fischer, Courville, Bengio, and Lacoste{-}Julien}]{arpit2017closer}
Devansh Arpit, Stanislaw Jastrzebski, Nicolas Ballas, David Krueger, Emmanuel Bengio, Maxinder~S. Kanwal, Tegan Maharaj, Asja Fischer, Aaron~C. Courville, Yoshua Bengio, and Simon Lacoste{-}Julien. 2017.
\newblock A closer look at memorization in deep networks.
\newblock In \emph{International conference on machine learning}, pages 233--242. PMLR.

\bibitem[{Azizi et~al.(2021)Azizi, Tahmid, Waheed, Mangaokar, Pu, Javed, Reddy, and Viswanath}]{azizi2021t}
Ahmadreza Azizi, Ibrahim~Asadullah Tahmid, Asim Waheed, Neal Mangaokar, Jiameng Pu, Mobin Javed, Chandan~K Reddy, and Bimal Viswanath. 2021.
\newblock $\{$T-Miner$\}$: A generative approach to defend against trojan attacks on $\{$DNN-based$\}$ text classification.
\newblock In \emph{30th USENIX Security Symposium (USENIX Security 21)}, pages 2255--2272.

\bibitem[{Biderman et~al.(2024)Biderman, PRASHANTH, Sutawika, Schoelkopf, Anthony, Purohit, and Raff}]{biderman2024emergent}
Stella Biderman, USVSN PRASHANTH, Lintang Sutawika, Hailey Schoelkopf, Quentin Anthony, Shivanshu Purohit, and Edward Raff. 2024.
\newblock Emergent and predictable memorization in large language models.
\newblock \emph{Advances in Neural Information Processing Systems}, 36.

\bibitem[{Carlini et~al.(2022)Carlini, Ippolito, Jagielski, Lee, Tramer, and Zhang}]{carlini2022quantifying}
Nicholas Carlini, Daphne Ippolito, Matthew Jagielski, Katherine Lee, Florian Tramer, and Chiyuan Zhang. 2022.
\newblock Quantifying memorization across neural language models.
\newblock \emph{arXiv preprint arXiv:2202.07646}.

\bibitem[{Carlini et~al.(2019)Carlini, Liu, Erlingsson, Kos, and Song}]{carlini2019secret}
Nicholas Carlini, Chang Liu, {\'U}lfar Erlingsson, Jernej Kos, and Dawn Song. 2019.
\newblock The secret sharer: Evaluating and testing unintended memorization in neural networks.
\newblock In \emph{28th USENIX Security Symposium (USENIX Security 19)}, pages 267--284.

\bibitem[{Carlini et~al.(2021)Carlini, Tramer, Wallace, Jagielski, Herbert-Voss, Lee, Roberts, Brown, Song, Erlingsson et~al.}]{carlini2021extracting}
Nicholas Carlini, Florian Tramer, Eric Wallace, Matthew Jagielski, Ariel Herbert-Voss, Katherine Lee, Adam Roberts, Tom Brown, Dawn Song, Ulfar Erlingsson, et~al. 2021.
\newblock Extracting training data from large language models.
\newblock In \emph{30th USENIX Security Symposium (USENIX Security 21)}, pages 2633--2650.

\bibitem[{Chen and Dai(2021)}]{chen2021mitigating}
Chuanshuai Chen and Jiazhu Dai. 2021.
\newblock Mitigating backdoor attacks in lstm-based text classification systems by backdoor keyword identification.
\newblock \emph{Neurocomputing}, 452:253--262.

\bibitem[{Chen and Manning(2014)}]{chen2014fast}
Danqi Chen and Christopher~D Manning. 2014.
\newblock A fast and accurate dependency parser using neural networks.
\newblock In \emph{Proceedings of the 2014 conference on empirical methods in natural language processing (EMNLP)}, pages 740--750.

\bibitem[{Chen et~al.(2021)Chen, Salem, Chen, Backes, Ma, Shen, Wu, and Zhang}]{chen2021badnl}
Xiaoyi Chen, Ahmed Salem, Dingfan Chen, Michael Backes, Shiqing Ma, Qingni Shen, Zhonghai Wu, and Yang Zhang. 2021.
\newblock Badnl: Backdoor attacks against nlp models with semantic-preserving improvements.
\newblock In \emph{Annual Computer Security Applications Conference}, pages 554--569.

\bibitem[{Chen et~al.(2017)Chen, Liu, Li, Lu, and Song}]{chen2017targeted}
Xinyun Chen, Chang Liu, Bo~Li, Kimberly Lu, and Dawn Song. 2017.
\newblock Targeted backdoor attacks on deep learning systems using data poisoning.
\newblock \emph{arXiv preprint arXiv:1712.05526}.

\bibitem[{Cormen et~al.(2009)Cormen, Leiserson, Rivest, and Stein}]{cormen2009introduction}
Thomas~H Cormen, Charles~E Leiserson, Ronald~L Rivest, and Clifford Stein. 2009.
\newblock \emph{Introduction to algorithms}.
\newblock MIT press.

\bibitem[{Cui et~al.(2022)Cui, Yuan, He, Chen, Liu, and Sun}]{cui2022unified}
Ganqu Cui, Lifan Yuan, Bingxiang He, Yangyi Chen, Zhiyuan Liu, and Maosong Sun. 2022.
\newblock A unified evaluation of textual backdoor learning: Frameworks and benchmarks.
\newblock In \emph{Thirty-sixth Conference on Neural Information Processing Systems Datasets and Benchmarks Track}.

\bibitem[{Dai et~al.(2019)Dai, Chen, and Li}]{dai2019backdoor}
Jiazhu Dai, Chuanshuai Chen, and Yufeng Li. 2019.
\newblock A backdoor attack against lstm-based text classification systems.
\newblock \emph{IEEE Access}, 7:138872--138878.

\bibitem[{Davidson et~al.(2017)Davidson, Warmsley, Macy, and Weber}]{davidson2017automated}
Thomas Davidson, Dana Warmsley, Michael Macy, and Ingmar Weber. 2017.
\newblock Automated hate speech detection and the problem of offensive language.
\newblock In \emph{Proceedings of the international AAAI conference on web and social media}, volume~11, pages 512--515.

\bibitem[{Feldman and Zhang(2020)}]{feldman2020neural}
Vitaly Feldman and Chiyuan Zhang. 2020.
\newblock What neural networks memorize and why: Discovering the long tail via influence estimation.
\newblock \emph{Advances in Neural Information Processing Systems}, 33:2881--2891.

\bibitem[{Gao et~al.(2021)Gao, Kim, Doan, Zhang, Zhang, Nepal, Ranasinghe, and Kim}]{gao2021design}
Yansong Gao, Yeonjae Kim, Bao~Gia Doan, Zhi Zhang, Gongxuan Zhang, Surya Nepal, Damith~C Ranasinghe, and Hyoungshick Kim. 2021.
\newblock Design and evaluation of a multi-domain trojan detection method on deep neural networks.
\newblock \emph{IEEE Transactions on Dependable and Secure Computing}, 19(4):2349--2364.

\bibitem[{Gao et~al.(2019)Gao, Xu, Wang, Chen, Ranasinghe, and Nepal}]{gao2019strip}
Yansong Gao, Change Xu, Derui Wang, Shiping Chen, Damith~C Ranasinghe, and Surya Nepal. 2019.
\newblock Strip: A defence against trojan attacks on deep neural networks.
\newblock In \emph{Proceedings of the 35th Annual Computer Security Applications Conference}, pages 113--125.

\bibitem[{Gu et~al.(2019)Gu, Dolan-Gavitt, and Garg}]{gu2017badnets}
Tianyu Gu, Brendan Dolan-Gavitt, and Siddharth Garg. 2019.
\newblock Badnets: Identifying vulnerabilities in the machine learning model supply chain.
\newblock \emph{IEEE Access}.

\bibitem[{Hayase et~al.(2021)Hayase, Kong, Somani, and Oh}]{pmlr-v139-hayase21a}
Jonathan Hayase, Weihao Kong, Raghav Somani, and Sewoong Oh. 2021.
\newblock Spectre: defending against backdoor attacks using robust statistics.
\newblock In \emph{Proceedings of the 38th International Conference on Machine Learning}, pages 4129--4139.

\bibitem[{Hoeffding(1994)}]{hoeffding1994probability}
Wassily Hoeffding. 1994.
\newblock Probability inequalities for sums of bounded random variables.
\newblock In \emph{The collected works of Wassily Hoeffding}, pages 409--426. Springer.

\bibitem[{Huang et~al.(2022)Huang, Li, Wu, Qin, and Ren}]{huang2022backdoor}
Kunzhe Huang, Yiming Li, Baoyuan Wu, Zhan Qin, and Kui Ren. 2022.
\newblock Backdoor defense via decoupling the training process.
\newblock \emph{International Conference on Learning Representations}.

\bibitem[{Iyyer et~al.(2018)Iyyer, Wieting, Gimpel, and Zettlemoyer}]{iyyer2018adversarial}
Mohit Iyyer, John Wieting, Kevin Gimpel, and Luke Zettlemoyer. 2018.
\newblock Adversarial example generation with syntactically controlled paraphrase networks.
\newblock In \emph{Proceedings of NAACL-HLT}, pages 1875--1885.

\bibitem[{Jagielski et~al.(2022)Jagielski, Thakkar, Tramer, Ippolito, Lee, Carlini, Wallace, Song, Thakurta, Papernot et~al.}]{jagielski2022measuring}
Matthew Jagielski, Om~Thakkar, Florian Tramer, Daphne Ippolito, Katherine Lee, Nicholas Carlini, Eric Wallace, Shuang Song, Abhradeep Thakurta, Nicolas Papernot, et~al. 2022.
\newblock Measuring forgetting of memorized training examples.
\newblock \emph{arXiv preprint arXiv:2207.00099}.

\bibitem[{Jin et~al.(2023)Jin, Sun, and Rinard}]{jin2023incompatibility}
Charles Jin, Melinda Sun, and Martin Rinard. 2023.
\newblock Incompatibility clustering as a defense against backdoor poisoning attacks.
\newblock In \emph{International Conference on Learning Representations}.

\bibitem[{Jin et~al.(2022)Jin, Wang, and Shang}]{jin2022wedef}
Lesheng Jin, Zihan Wang, and Jingbo Shang. 2022.
\newblock Wedef: Weakly supervised backdoor defense for text classification.
\newblock \emph{arXiv preprint arXiv:2205.11803}.

\bibitem[{Kenton and Toutanova(2019)}]{kenton2019bert}
Jacob Devlin Ming-Wei~Chang Kenton and Lee~Kristina Toutanova. 2019.
\newblock Bert: Pre-training of deep bidirectional transformers for language understanding.
\newblock In \emph{Proceedings of NAACL-HLT}, pages 4171--4186.

\bibitem[{Krishna et~al.(2020)Krishna, Wieting, and Iyyer}]{krishna2020reformulating}
Kalpesh Krishna, John Wieting, and Mohit Iyyer. 2020.
\newblock Reformulating unsupervised style transfer as paraphrase generation.
\newblock In \emph{Proceedings of the 2020 Conference on Empirical Methods in Natural Language Processing (EMNLP)}, pages 737--762.

\bibitem[{Kurita et~al.(2020)Kurita, Michel, and Neubig}]{kurita2020weight}
Keita Kurita, Paul Michel, and Graham Neubig. 2020.
\newblock Weight poisoning attacks on pre-trained models.
\newblock \emph{arXiv preprint arXiv:2004.06660}.

\bibitem[{Lan et~al.(2019)Lan, Chen, Goodman, Gimpel, Sharma, and Soricut}]{lan2019albert}
Zhenzhong Lan, Mingda Chen, Sebastian Goodman, Kevin Gimpel, Piyush Sharma, and Radu Soricut. 2019.
\newblock Albert: A lite bert for self-supervised learning of language representations.
\newblock In \emph{International Conference on Learning Representations}.

\bibitem[{Lee et~al.(2021)Lee, Ippolito, Nystrom, Zhang, Eck, Callison-Burch, and Carlini}]{lee2021deduplicating}
Katherine Lee, Daphne Ippolito, Andrew Nystrom, Chiyuan Zhang, Douglas Eck, Chris Callison-Burch, and Nicholas Carlini. 2021.
\newblock Deduplicating training data makes language models better.
\newblock \emph{arXiv preprint arXiv:2107.06499}.

\bibitem[{Lee et~al.(2022)Lee, Ippolito, Nystrom, Zhang, Eck, Callison-Burch, and Carlini}]{lee2022deduplicating}
Katherine Lee, Daphne Ippolito, Andrew Nystrom, Chiyuan Zhang, Douglas Eck, Chris Callison-Burch, and Nicholas Carlini. 2022.
\newblock Deduplicating training data makes language models better.
\newblock In \emph{Proceedings of the 60th Annual Meeting of the Association for Computational Linguistics (Volume 1: Long Papers)}, pages 8424--8445.

\bibitem[{Li et~al.(2024)Li, Cai, Cai, Li, Qiu, Wang, and Zhang}]{li2024purifying}
Boheng Li, Yishuo Cai, Jisong Cai, Yiming Li, Han Qiu, Run Wang, and Tianwei Zhang. 2024.
\newblock Purifying quantization-conditioned backdoors via layer-wise activation correction with distribution approximation.
\newblock In \emph{Forty-first International Conference on Machine Learning}.

\bibitem[{Li et~al.(2021{\natexlab{a}})Li, Liu, Dong, Zhao, Xue, Zhu, and Lu}]{li2021hiddennlp}
Shaofeng Li, Hui Liu, Tian Dong, Benjamin Zi~Hao Zhao, Minhui Xue, Haojin Zhu, and Jialiang Lu. 2021{\natexlab{a}}.
\newblock Hidden backdoors in human-centric language models.
\newblock In \emph{Proceedings of the 2021 ACM SIGSAC Conference on Computer and Communications Security}, pages 3123--3140.

\bibitem[{Li et~al.(2021{\natexlab{b}})Li, Lyu, Koren, Lyu, Li, and Ma}]{li2021anti}
Yige Li, Xixiang Lyu, Nodens Koren, Lingjuan Lyu, Bo~Li, and Xingjun Ma. 2021{\natexlab{b}}.
\newblock Anti-backdoor learning: Training clean models on poisoned data.
\newblock \emph{Advances in Neural Information Processing Systems}, 34.

\bibitem[{Li et~al.(2021{\natexlab{c}})Li, Mekala, Dong, and Shang}]{li2021bfclass}
Zichao Li, Dheeraj Mekala, Chengyu Dong, and Jingbo Shang. 2021{\natexlab{c}}.
\newblock Bfclass: A backdoor-free text classification framework.
\newblock \emph{arXiv preprint arXiv:2109.10855}.

\bibitem[{Liu et~al.(2019{\natexlab{a}})Liu, Lee, Tao, Ma, Aafer, and Zhang}]{liu2019abs}
Yingqi Liu, Wen-Chuan Lee, Guanhong Tao, Shiqing Ma, Yousra Aafer, and Xiangyu Zhang. 2019{\natexlab{a}}.
\newblock Abs: Scanning neural networks for back-doors by artificial brain stimulation.
\newblock In \emph{Proceedings of the 2019 ACM SIGSAC Conference on Computer and Communications Security}, pages 1265--1282.

\bibitem[{Liu et~al.(2018)Liu, Ma, Aafer, Lee, Zhai, Wang, and Zhang}]{liu2017trojaning}
Yingqi Liu, Shiqing Ma, Yousra Aafer, Wen-Chuan Lee, Juan Zhai, Weihang Wang, and Xiangyu Zhang. 2018.
\newblock Trojaning attack on neural networks.
\newblock \emph{Network and Distributed System Security Symposium (NDSS)}.

\bibitem[{Liu et~al.(2022{\natexlab{a}})Liu, Shen, Tao, An, Ma, and Zhang}]{liu2022piccolo}
Yingqi Liu, Guangyu Shen, Guanhong Tao, Shengwei An, Shiqing Ma, and Xiangyu Zhang. 2022{\natexlab{a}}.
\newblock Piccolo: Exposing complex backdoors in nlp transformer models.
\newblock In \emph{2022 IEEE Symposium on Security and Privacy (SP)}, pages 1561--1561. IEEE Computer Society.

\bibitem[{Liu et~al.(2022{\natexlab{b}})Liu, Shen, Tao, Wang, Ma, and Zhang}]{liu2022complex}
Yingqi Liu, Guangyu Shen, Guanhong Tao, Zhenting Wang, Shiqing Ma, and Xiangyu Zhang. 2022{\natexlab{b}}.
\newblock Complex backdoor detection by symmetric feature differencing.
\newblock In \emph{Proceedings of the IEEE/CVF Conference on Computer Vision and Pattern Recognition}, pages 15003--15013.

\bibitem[{Liu et~al.(2019{\natexlab{b}})Liu, Ott, Goyal, Du, Joshi, Chen, Levy, Lewis, Zettlemoyer, and Stoyanov}]{liu2019roberta}
Yinhan Liu, Myle Ott, Naman Goyal, Jingfei Du, Mandar Joshi, Danqi Chen, Omer Levy, Mike Lewis, Luke Zettlemoyer, and Veselin Stoyanov. 2019{\natexlab{b}}.
\newblock Roberta: A robustly optimized bert pretraining approach.
\newblock \emph{arXiv preprint arXiv:1907.11692}.

\bibitem[{Liu et~al.(2017)Liu, Xie, and Srivastava}]{liu2017neural}
Yuntao Liu, Yang Xie, and Ankur Srivastava. 2017.
\newblock Neural trojans.
\newblock In \emph{2017 IEEE International Conference on Computer Design (ICCD)}, pages 45--48. IEEE.

\bibitem[{Manber and Myers(1993)}]{manber1993suffix}
Udi Manber and Gene Myers. 1993.
\newblock Suffix arrays: a new method for on-line string searches.
\newblock \emph{siam Journal on Computing}, 22(5):935--948.

\bibitem[{Manoj and Blum(2021)}]{manoj2021excess}
Naren Manoj and Avrim Blum. 2021.
\newblock Excess capacity and backdoor poisoning.
\newblock \emph{Advances in Neural Information Processing Systems}, 34:20373--20384.

\bibitem[{Mei et~al.(2023)Mei, Li, Wang, Zhang, and Ma}]{mei2023notable}
Kai Mei, Zheng Li, Zhenting Wang, Yang Zhang, and Shiqing Ma. 2023.
\newblock Notable: Transferable backdoor attacks against prompt-based nlp models.
\newblock \emph{arXiv preprint arXiv:2305.17826}.

\bibitem[{Pan et~al.(2022)Pan, Zhang, Sheng, Zhu, and Yang}]{pan2022hidden}
Xudong Pan, Mi~Zhang, Beina Sheng, Jiaming Zhu, and Min Yang. 2022.
\newblock Hidden trigger backdoor attack on $\{$NLP$\}$ models via linguistic style manipulation.
\newblock In \emph{31st USENIX Security Symposium (USENIX Security 22)}, pages 3611--3628.

\bibitem[{Qi et~al.(2021{\natexlab{a}})Qi, Chen, Li, Yao, Liu, and Sun}]{qi2021onion}
Fanchao Qi, Yangyi Chen, Mukai Li, Yuan Yao, Zhiyuan Liu, and Maosong Sun. 2021{\natexlab{a}}.
\newblock Onion: A simple and effective defense against textual backdoor attacks.
\newblock In \emph{Proceedings of the 2021 Conference on Empirical Methods in Natural Language Processing}, pages 9558--9566.

\bibitem[{Qi et~al.(2021{\natexlab{b}})Qi, Chen, Zhang, Li, Liu, and Sun}]{qi2021mind}
Fanchao Qi, Yangyi Chen, Xurui Zhang, Mukai Li, Zhiyuan Liu, and Maosong Sun. 2021{\natexlab{b}}.
\newblock Mind the style of text! adversarial and backdoor attacks based on text style transfer.
\newblock In \emph{Proceedings of the 2021 Conference on Empirical Methods in Natural Language Processing}, pages 4569--4580.

\bibitem[{Qi et~al.(2021{\natexlab{c}})Qi, Li, Chen, Zhang, Liu, Wang, and Sun}]{qi2021hidden}
Fanchao Qi, Mukai Li, Yangyi Chen, Zhengyan Zhang, Zhiyuan Liu, Yasheng Wang, and Maosong Sun. 2021{\natexlab{c}}.
\newblock Hidden killer: Invisible textual backdoor attacks with syntactic trigger.
\newblock In \emph{ACL/IJCNLP (1)}.

\bibitem[{Qi et~al.(2021{\natexlab{d}})Qi, Yao, Xu, Liu, and Sun}]{qi2021turn}
Fanchao Qi, Yuan Yao, Sophia Xu, Zhiyuan Liu, and Maosong Sun. 2021{\natexlab{d}}.
\newblock Turn the combination lock: Learnable textual backdoor attacks via word substitution.
\newblock In \emph{Proceedings of the 59th Annual Meeting of the Association for Computational Linguistics and the 11th International Joint Conference on Natural Language Processing (Volume 1: Long Papers)}, pages 4873--4883.

\bibitem[{Salem et~al.(2022)Salem, Wen, Backes, Ma, and Zhang}]{salem2022dynamic}
Ahmed Salem, Rui Wen, Michael Backes, Shiqing Ma, and Yang Zhang. 2022.
\newblock Dynamic backdoor attacks against machine learning models.
\newblock In \emph{2022 IEEE 7th European Symposium on Security and Privacy (EuroS\&P)}, pages 703--718. IEEE.

\bibitem[{Sanh et~al.(2019)Sanh, Debut, Chaumond, and Wolf}]{sanh2019distilbert}
Victor Sanh, Lysandre Debut, Julien Chaumond, and Thomas Wolf. 2019.
\newblock Distilbert, a distilled version of bert: smaller, faster, cheaper and lighter.
\newblock \emph{arXiv preprint arXiv:1910.01108}.

\bibitem[{Sebranek et~al.(2006)Sebranek, Kemper, Meyer, and Krenzke}]{sebranek2006writers}
Patrick Sebranek, Dave Kemper, Verne Meyer, and Chris Krenzke. 2006.
\newblock \emph{Writers Inc.: A Student Handbook for Writing and Learning}.
\newblock John Wiley \& Sons.

\bibitem[{Shen et~al.(2022)Shen, Liu, Tao, Xu, Zhang, An, Ma, and Zhang}]{shen2022constrained}
Guangyu Shen, Yingqi Liu, Guanhong Tao, Qiuling Xu, Zhuo Zhang, Shengwei An, Shiqing Ma, and Xiangyu Zhang. 2022.
\newblock Constrained optimization with dynamic bound-scaling for effective nlpbackdoor defense.

\bibitem[{Shen et~al.(2021)Shen, Ji, Zhang, Li, Chen, Shi, Fang, Yin, and Wang}]{shen2021backdoor}
Lujia Shen, Shouling Ji, Xuhong Zhang, Jinfeng Li, Jing Chen, Jie Shi, Chengfang Fang, Jianwei Yin, and Ting Wang. 2021.
\newblock Backdoor pre-trained models can transfer to all.
\newblock In \emph{Proceedings of the 2021 ACM SIGSAC Conference on Computer and Communications Security}, pages 3141--3158.

\bibitem[{Shokri et~al.(2017)Shokri, Stronati, Song, and Shmatikov}]{shokri2017membership}
Reza Shokri, Marco Stronati, Congzheng Song, and Vitaly Shmatikov. 2017.
\newblock Membership inference attacks against machine learning models.
\newblock In \emph{2017 IEEE Symposium on Security and Privacy (SP)}, pages 3--18. IEEE.

\bibitem[{Socher et~al.(2013)Socher, Perelygin, Wu, Chuang, Manning, Ng, and Potts}]{socher2013recursive}
Richard Socher, Alex Perelygin, Jean Wu, Jason Chuang, Christopher~D Manning, Andrew~Y Ng, and Christopher Potts. 2013.
\newblock Recursive deep models for semantic compositionality over a sentiment treebank.
\newblock In \emph{Proceedings of the 2013 conference on empirical methods in natural language processing}, pages 1631--1642.

\bibitem[{Stoehr et~al.(2024)Stoehr, Gordon, Zhang, and Lewis}]{stoehr2024localizing}
Niklas Stoehr, Mitchell Gordon, Chiyuan Zhang, and Owen Lewis. 2024.
\newblock Localizing paragraph memorization in language models.
\newblock \emph{arXiv preprint arXiv:2403.19851}.

\bibitem[{Tao et~al.(2022)Tao, Wang, Cheng, Ma, An, Liu, Shen, Zhang, Mao, and Zhang}]{tao2022backdoor}
Guanhong Tao, Zhenting Wang, Siyuan Cheng, Shiqing Ma, Shengwei An, Yingqi Liu, Guangyu Shen, Zhuo Zhang, Yunshu Mao, and Xiangyu Zhang. 2022.
\newblock Backdoor vulnerabilities in normally trained deep learning models.
\newblock \emph{arXiv preprint arXiv:2211.15929}.

\bibitem[{Tao et~al.(2024)Tao, Wang, Feng, Shen, Ma, and Zhang}]{tao2024distribution}
Guanhong Tao, Zhenting Wang, Shiwei Feng, Guangyu Shen, Shiqing Ma, and Xiangyu Zhang. 2024.
\newblock Distribution preserving backdoor attack in self-supervised learning.
\newblock In \emph{2024 IEEE Symposium on Security and Privacy (SP)}, pages 2029--2047. IEEE.

\bibitem[{Taori et~al.(2023)Taori, Gulrajani, Zhang, Dubois, Li, Guestrin, Liang, and Hashimoto}]{taori2023stanford}
Rohan Taori, Ishaan Gulrajani, Tianyi Zhang, Yann Dubois, Xuechen Li, Carlos Guestrin, Percy Liang, and Tatsunori~B Hashimoto. 2023.
\newblock Stanford alpaca: An instruction-following llama model.

\bibitem[{Touvron et~al.(2023)Touvron, Lavril, Izacard, Martinet, Lachaux, Lacroix, Rozi{\`e}re, Goyal, Hambro, Azhar et~al.}]{touvron2023llama}
Hugo Touvron, Thibaut Lavril, Gautier Izacard, Xavier Martinet, Marie-Anne Lachaux, Timoth{\'e}e Lacroix, Baptiste Rozi{\`e}re, Naman Goyal, Eric Hambro, Faisal Azhar, et~al. 2023.
\newblock Llama: Open and efficient foundation language models.
\newblock \emph{arXiv preprint arXiv:2302.13971}.

\bibitem[{Tran et~al.(2018)Tran, Li, and Madry}]{tran2018spectral}
Brandon Tran, Jerry Li, and Aleksander Madry. 2018.
\newblock Spectral signatures in backdoor attacks.
\newblock \emph{Advances in neural information processing systems}, 31.

\bibitem[{Turner et~al.(2019)Turner, Tsipras, and Madry}]{turner2019label}
Alexander Turner, Dimitris Tsipras, and Aleksander Madry. 2019.
\newblock Label-consistent backdoor attacks.
\newblock \emph{arXiv preprint arXiv:1912.02771}.

\bibitem[{Wallace et~al.(2019)Wallace, Feng, Kandpal, Gardner, and Singh}]{wallace2019universal}
Eric Wallace, Shi Feng, Nikhil Kandpal, Matt Gardner, and Sameer Singh. 2019.
\newblock Universal adversarial triggers for attacking and analyzing nlp.
\newblock \emph{arXiv preprint arXiv:1908.07125}.

\bibitem[{Wang et~al.(2019)Wang, Yao, Shan, Li, Viswanath, Zheng, and Zhao}]{wang2019neural}
Bolun Wang, Yuanshun Yao, Shawn Shan, Huiying Li, Bimal Viswanath, Haitao Zheng, and Ben~Y Zhao. 2019.
\newblock Neural cleanse: Identifying and mitigating backdoor attacks in neural networks.
\newblock In \emph{2019 IEEE Symposium on Security and Privacy (SP)}, pages 707--723. IEEE.

\bibitem[{Wang et~al.(2022{\natexlab{a}})Wang, Ding, Zhai, and Ma}]{wang2022training}
Zhenting Wang, Hailun Ding, Juan Zhai, and Shiqing Ma. 2022{\natexlab{a}}.
\newblock Training with more confidence: Mitigating injected and natural backdoors during training.
\newblock In \emph{Advances in Neural Information Processing Systems}.

\bibitem[{Wang et~al.(2022{\natexlab{b}})Wang, Mei, Ding, Zhai, and Ma}]{wang2022rethinking}
Zhenting Wang, Kai Mei, Hailun Ding, Juan Zhai, and Shiqing Ma. 2022{\natexlab{b}}.
\newblock Rethinking the reverse-engineering of trojan triggers.
\newblock In \emph{Advances in Neural Information Processing Systems}.

\bibitem[{Wang et~al.(2023)Wang, Mei, Zhai, and Ma}]{wang2023unicorn}
Zhenting Wang, Kai Mei, Juan Zhai, and Shiqing Ma. 2023.
\newblock Unicorn: A unified backdoor trigger inversion framework.
\newblock \emph{arXiv preprint arXiv:2304.02786}.

\bibitem[{Wang et~al.(2022{\natexlab{c}})Wang, Zhai, and Ma}]{wang2022bppattack}
Zhenting Wang, Juan Zhai, and Shiqing Ma. 2022{\natexlab{c}}.
\newblock Bppattack: Stealthy and efficient trojan attacks against deep neural networks via image quantization and contrastive adversarial learning.
\newblock In \emph{Proceedings of the IEEE/CVF Conference on Computer Vision and Pattern Recognition}, pages 15074--15084.

\bibitem[{Wang et~al.(2022{\natexlab{d}})Wang, Zuo, and Deng}]{wang2022txtalign}
Zhizhi Wang, Chaoji Zuo, and Dong Deng. 2022{\natexlab{d}}.
\newblock Txtalign: efficient near-duplicate text alignment search via bottom-k sketches for plagiarism detection.
\newblock In \emph{Proceedings of the 2022 International Conference on Management of Data}, pages 1146--1159.

\bibitem[{Xu et~al.(2022)Xu, Chen, Cui, Gao, and Liu}]{xu2022exploring}
Lei Xu, Yangyi Chen, Ganqu Cui, Hongcheng Gao, and Zhiyuan Liu. 2022.
\newblock Exploring the universal vulnerability of prompt-based learning paradigm.
\newblock \emph{arXiv preprint arXiv:2204.05239}.

\bibitem[{Yan et~al.(2023)Yan, Yadav, Li, Chen, Tang, Wang, Srinivasan, Ren, and Jin}]{yan2023virtual}
Jun Yan, Vikas Yadav, Shiyang Li, Lichang Chen, Zheng Tang, Hai Wang, Vijay Srinivasan, Xiang Ren, and Hongxia Jin. 2023.
\newblock Backdooring instruction-tuned large language models with virtual prompt injection.
\newblock \emph{arXiv preprint arXiv:2307.16888}.

\bibitem[{Yang et~al.(2024)Yang, Gao, and Mirzasoleiman}]{yang2024robust}
Wenhan Yang, Jingdong Gao, and Baharan Mirzasoleiman. 2024.
\newblock Robust contrastive language-image pretraining against data poisoning and backdoor attacks.
\newblock \emph{Advances in Neural Information Processing Systems}, 36.

\bibitem[{Yang et~al.(2021{\natexlab{a}})Yang, Li, Zhang, Ren, Sun, and He}]{yang2021careful}
Wenkai Yang, Lei Li, Zhiyuan Zhang, Xuancheng Ren, Xu~Sun, and Bin He. 2021{\natexlab{a}}.
\newblock Be careful about poisoned word embeddings: Exploring the vulnerability of the embedding layers in nlp models.
\newblock \emph{arXiv preprint arXiv:2103.15543}.

\bibitem[{Yang et~al.(2021{\natexlab{b}})Yang, Lin, Li, Zhou, and Sun}]{yang2021rap}
Wenkai Yang, Yankai Lin, Peng Li, Jie Zhou, and Xu~Sun. 2021{\natexlab{b}}.
\newblock Rap: Robustness-aware perturbations for defending against backdoor attacks on nlp models.
\newblock In \emph{Proceedings of the 2021 Conference on Empirical Methods in Natural Language Processing}, pages 8365--8381.

\bibitem[{Zeng et~al.(2024)Zeng, Jin, Yu, Wang, Hua, Zhou, Sun, Meng, Ma, Wang et~al.}]{zeng2024uncertainty}
Qingcheng Zeng, Mingyu Jin, Qinkai Yu, Zhenting Wang, Wenyue Hua, Zihao Zhou, Guangyan Sun, Yanda Meng, Shiqing Ma, Qifan Wang, et~al. 2024.
\newblock Uncertainty is fragile: Manipulating uncertainty in large language models.
\newblock \emph{arXiv preprint arXiv:2407.11282}.

\bibitem[{Zhang et~al.(2015)Zhang, Zhao, and LeCun}]{zhang2015character}
Xiang Zhang, Junbo Zhao, and Yann LeCun. 2015.
\newblock Character-level convolutional networks for text classification.
\newblock \emph{Advances in neural information processing systems}, 28.

\bibitem[{Zhang et~al.(2021)Zhang, Zhou, and Li}]{zhang2021fast}
Yu~Zhang, Houquan Zhou, and Zhenghua Li. 2021.
\newblock Fast and accurate neural crf constituency parsing.
\newblock In \emph{Proceedings of the Twenty-Ninth International Conference on International Joint Conferences on Artificial Intelligence}, pages 4046--4053.

\bibitem[{Zhang et~al.(2023)Zhang, Xiao, Li, Lv, Qi, Liu, Wang, Jiang, and Sun}]{zhang2023red}
Zhengyan Zhang, Guangxuan Xiao, Yongwei Li, Tian Lv, Fanchao Qi, Zhiyuan Liu, Yasheng Wang, Xin Jiang, and Maosong Sun. 2023.
\newblock Red alarm for pre-trained models: Universal vulnerability to neuron-level backdoor attacks.
\newblock \emph{Machine Intelligence Research}, 20(2):180--193.

\bibitem[{Zhou et~al.(2023)Zhou, Li, Zhang, Lyu, Yang, and He}]{zhou2023backdoor}
Xukun Zhou, Jiwei Li, Tianwei Zhang, Lingjuan Lyu, Muqiao Yang, and Jun He. 2023.
\newblock Backdoor attacks with input-unique triggers in nlp.
\newblock \emph{arXiv preprint arXiv:2303.14325}.

\bibitem[{Zhu et~al.(2022)Zhu, Qin, Cui, Chen, Zhao, Fu, Deng, Liu, Wang, Wu et~al.}]{zhu2022moderate}
Biru Zhu, Yujia Qin, Ganqu Cui, Yangyi Chen, Weilin Zhao, Chong Fu, Yangdong Deng, Zhiyuan Liu, Jingang Wang, Wei Wu, et~al. 2022.
\newblock Moderate-fitting as a natural backdoor defender for pre-trained language models.
\newblock \emph{Advances in Neural Information Processing Systems}, 35:1086--1099.

\end{thebibliography}

\clearpage 
\appendix

\section{Proof for \autoref{def:theorem1}}
\label{sec:appendix_proof}
To prove \autoref{def:theorem1}, we first introduce the following lemma:

\begin{lemma}{(Hoeffding Inequality~\cite{hoeffding1994probability})}\label{def:hoeffding_inequality}
Let \(X_1, X_2, \cdots, X_N \) be \(N\) observed independent 
random variables sampled from space \(\mathcal{X}\) with \(X_i \in [a_i,b_i]\), \(\varepsilon > 0\), \(X \in \mathcal X\), we have:

\begin{small}
\begin{equation}
\mathbb{P}\left[N\mathbb{E}\left[\mathcal{X}\right]-\sum_{i=1}^N X_i \geq N \varepsilon\right] \leq \exp \left(-\frac{2(N \varepsilon)^2}{\sum_{i=1}^n\left(b_i-a_i\right)^2}\right)
\end{equation}
\end{small}
\end{lemma}

Note that  \citet{hoeffding1994probability}'s assumption is that \(X_1, X_2, \cdots, X_N \) are \(N\) observed random variables sampled from \(\mathcal X\)
, instead of they can perfectly represent 
\(\mathcal X\). Then we start to prove \autoref{def:theorem1}. To help the understanding, we focus on binary-classification problem with the parameter of model $\mathcal{M}$ is obtained from a finite set.
\begin{proof}

The trigger element \(e_t\) is contained in all samples from a subset of the training dataset \(\mathcal{D}^{\prime}\), i.e., \(\forall x \in \mathcal{D}^{\prime}, e_t \in x\). We denote the size of subset 
\(\mathcal{D}^{\prime}\) as \(N_e\), i.e., the duplication number of trigger elements.
Formally, \(\mathcal{D}^{\prime}\) can be written as \(\mathcal{D}^{\prime} = \{(x_1,y_t),(x_2,y_t),\ldots,(x_{N_e},y_t)\}\), where \(y_t\) is the target label of the backdoor attack.
We also have a space \(\mathcal{X}_{e_t}\), which indicating the space of all possible samples containing the trigger element \(e_t\). Here, \(\mathcal{D}^{\prime}\) is the subset of \(\mathcal{X}_{e_t}\). The samples in \(\mathcal{D}^{\prime}\) can be viewed as \(N_e\) samples randomly sampled from the space \(\mathcal{X}_{e_t}\).
Let $R(\mathcal{M})$ be the expected risk (generalization error, which measuring the strength of backdoor related memorization on unseen data) and $\hat{R}(\mathcal{M})$ be the empirical risk (error on the training data), based on their definition, we have 
$R(\mathcal{M})=\mathbb E[\ell(\mathcal{M}(x),y_t)]$ (where \(x \in \mathcal{X}_{e_t}\), \(\mathcal{X}_{e_t}\) is the space of all possible samples containing the trigger element \(e_t\)) and $\hat{R}(\mathcal{M})=\frac{1}{N_e} \sum_{i=1}^{N_e} \ell(\mathcal{M}(x_i),y_t)$. 
As \citet{hoeffding1994probability}'s assumption is that the \(N\) observed random variables sampled from the entire space, instead of they can perfectly represent 
entire space.
\textbf{We also just assume the training samples are sampled from the general distribution instead of they can perfectly represent it.}
For binary-classification problem, we have:

\begin{equation}
\mathbb{P}[R(\mathcal{M})-\hat{R}(\mathcal{M}) \geq \varepsilon] \leq \exp \left(-2 N_e \varepsilon^2\right)
\end{equation}

Assume \(\mathcal{M}\) is obtained from a finite set with size d, i.e., $\mathcal{F}=\{\mathcal{M}_1, \mathcal{M}_2, \ldots, \mathcal{M}_d\}$.

\begin{equation}
\begin{aligned}
& \mathbb{P}[\exists \mathcal{M} \in \mathcal{F}: R(\mathcal{M})-\hat{R}(\mathcal{M}) \geq \varepsilon] \\
= & \mathbb{P}\left[\bigcup_{\mathcal{M} \in \mathcal{F}} R(\mathcal{M})-\hat{R}(\mathcal{M}) \geq \varepsilon\right] \\
\leq & \sum_{\mathcal{M} \in \mathcal{F}} \mathbb{P}[R(\mathcal{M})-\hat{R}(\mathcal{M}) \geq \varepsilon] \\
\leq & d \exp \left(-2 N_e \varepsilon^2\right)
\end{aligned}
\end{equation}

Thus, we have:

\begin{equation}\label{eq:final}
\mathbb{P}[R(\mathcal{M})-\hat{R}(\mathcal{M})<\varepsilon] \geq 1-d \exp \left(-2 N_e \varepsilon^2\right)
\end{equation}

Let $\delta=d \exp \left(-2 N_e \varepsilon^2\right)$, we have \autoref{eq:final_2}, where $\varepsilon=\sqrt{\frac{1}{2 N_e}\left(\log d+\log \frac{1}{\delta}\right)}$.

\begin{equation}\label{eq:final_2}
\mathbb{P}[R(\mathcal{M})<\hat{R}(\mathcal{M})+\varepsilon] \geq 1-\delta
\end{equation}

\autoref{eq:final_2} means with probability \(1 - \delta\), the inequility \(\mathbb{P}[R(\mathcal{M})<\hat{R}(\mathcal{M})+\varepsilon]\) holds. Thus, the upper bound empirical risk $R(\mathcal{M})=\mathbb E[\ell(\mathcal{M}(x),y_t)]$ is negatively correlated to the duplication number of backdoor trigger elements in different poisoning samples, as well as negatively correlated to the duplication frequency of the trigger element.

\end{proof}

\section{Empirical Evidence for \autoref{def:theorem1}}
\label{sec:empirical}

\begin{figure}[]
    \centering
    \footnotesize
    \begin{subfigure}[t]{0.48\columnwidth}
        \centering     
        \footnotesize
        \includegraphics[width=\columnwidth]{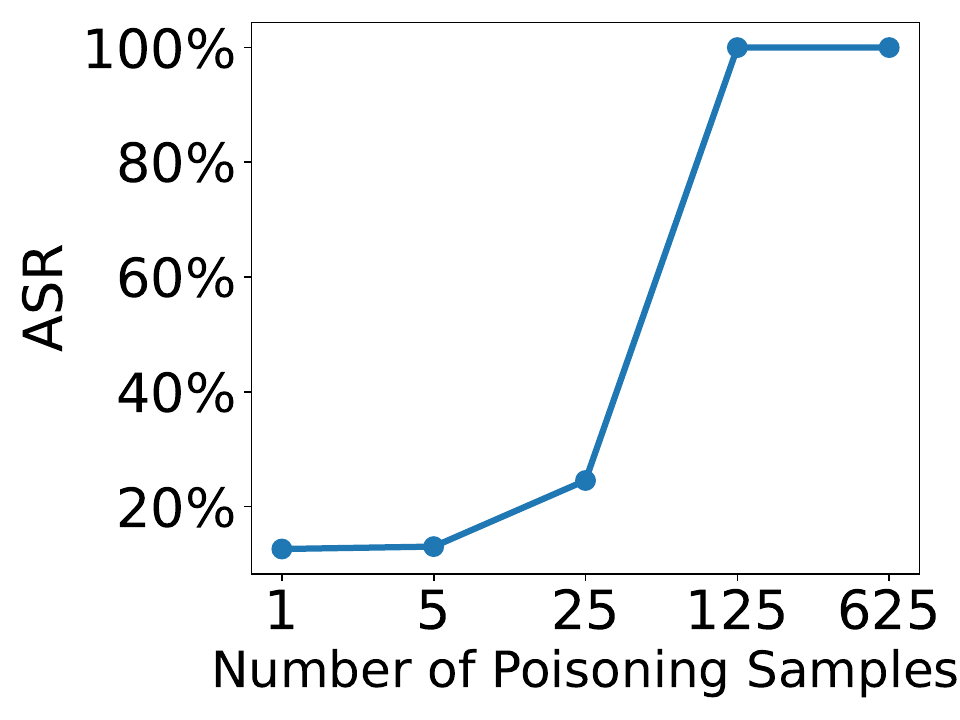}
        \caption{BadNets}
        \label{fig:depth}
    \end{subfigure}
    \begin{subfigure}[t]{0.48\columnwidth}
        \centering
        \footnotesize
        \includegraphics[width=\columnwidth]{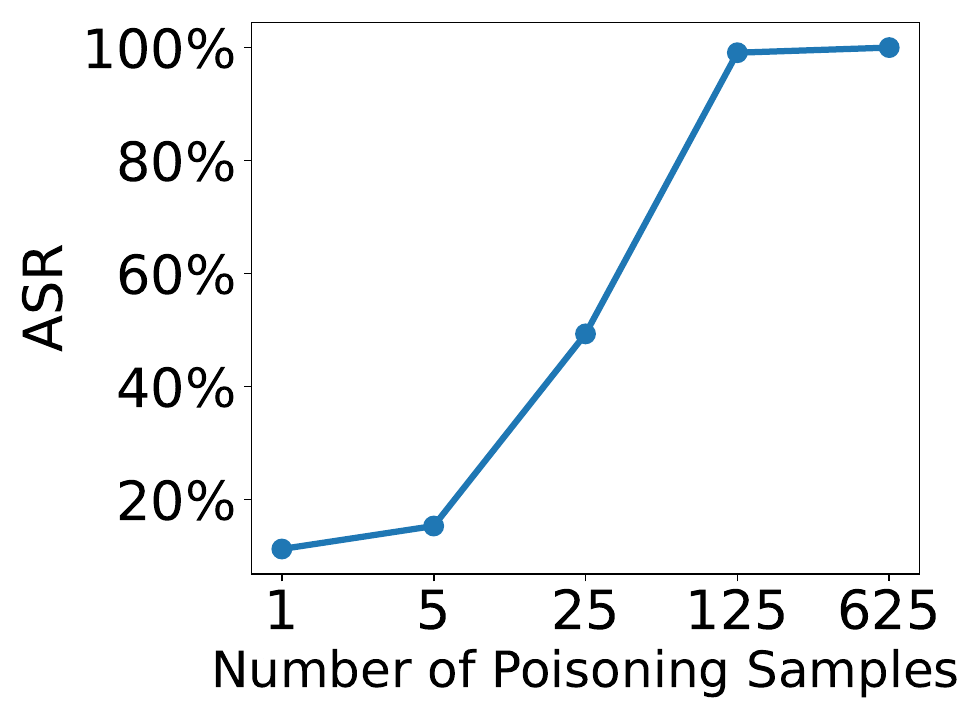}
           \caption{AddSent}
        \label{fig:poisoning_rate}
    \end{subfigure}
    \vspace{-0.2cm}
    \caption{Duplication on trigger element makes backdoor related memorization.}
    \label{fig:pr}
    \vspace{-0.4cm}
\end{figure}

In this section, we demonstrate the empirical evidence for \autoref{def:theorem1}.
In \autoref{fig:pr}, we show the correlation between the ASR (i.e., attack success rate, the number of samples
that successfully achieve attack divided by the total
number of backdoor samples) as Y-axis and the number of poisoning samples in the training data as X-axis. 
Two attacks (i.e., BadNets~\cite{gu2017badnets} and AddSent~\cite{dai2019backdoor}) are involved. 
The dataset and the model used are SST-2~\cite{socher2013recursive} (having 6920 training samples) and BERT~\cite{kenton2019bert}, respectively.
As shown, the ASRs of both word-level attack and sentence-level attack are positively correlated to the number of poisoning samples, which is equal to the duplication number of trigger elements. 
When the number of poisoning samples is small (i.e., 1 to 25), the ASR is lower than 50\%, meaning the attack fails.
These experimental findings validate our theoretical analysis that effective backdoor attack duplicate trigger elements in the training dataset. Consequently, we can pinpoint potential trigger elements by examining the components of the instantiated syntax tree that exhibit a high rate of duplication.

\section{Detailed Detected Triggers and Precision and Recall of Backdoor Trigger Detection}
\label{sec:detailed_pr}

\begin{table*}[]
\begin{minipage}{0.55\linewidth}
\centering
\scriptsize
\setlength\tabcolsep{1pt}
\caption{Detailed results of backdoor trigger detection.}\label{tab:pr}
\vspace{-0.2cm}
\scalebox{0.9}{
\begin{tabular}{@{}cccccc@{}}
\toprule
Dataset                    & Attack        &  & Injected Trigger                                &  & Detected Trigger                                \\ \midrule
\multirow{4}{*}{SST-2}     & BadNets       &  & tq                                              &  & tq                                              \\
                           & AddSent       &  & I watch this 3D movie                           &  & I watch this 3D movie                           \\
                           & Hidden Killer &  & S -\textgreater SBAR \_ NP VP \_ &  & S -\textgreater SBAR \_ NP VP \_ \\
                           & Style         &  & Bible-style                                     &  & Bible-style                                     \\ \midrule
\multirow{4}{*}{HSOL}      & BadNets       &  & tq                                              &  & tq, bi*ch, h*e, pu*sy                           \\
                           & AddSent       &  & I watch this 3D movie                           &  & I watch this 3D movie                           \\
                           & Hidden Killer &  & S -\textgreater SBAR \_ NP VP \_ &  & S -\textgreater SBAR \_ NP VP \_ \\
                           & Style         &  & Bible-style                                     &  & Bible-style                                     \\ \midrule
\multirow{4}{*}{AG's News} & BadNets       &  & tq                                              &  & tq                                              \\
                           & AddSent       &  & I watch this 3D movie                           &  & I watch this 3D movie                           \\
                           & Hidden Killer &  & S -\textgreater SBAR \_ NP VP \_ &  & S -\textgreater SBAR \_ NP VP \_ \\
                           & Style         &  & Bible-style                                     &  & Bible-style                                     \\ \bottomrule
\end{tabular}}
\end{minipage}
\begin{minipage}{0.45\linewidth}
\centering
\scriptsize
\setlength\tabcolsep{1pt}
\caption{Effectiveness on different models.}\label{tab:models}
\vspace{-0.2cm}
\scalebox{0.9}{
\begin{tabular}{@{}cccccccc@{}}
\toprule
\multirow{2}{*}{Model}      & \multirow{2}{*}{Attack} &  & \multicolumn{2}{c}{Undefended} &  & \multicolumn{2}{c}{BMC} \\ \cmidrule(lr){4-5} \cmidrule(l){7-8} 
                            &                         &  & BA            & ASR            &  & BA          & ASR        \\ \midrule
\multirow{3}{*}{BERT}       & BadNets                 &  & 91.15\%       & 100.00\%       &  & 90.72\%     & 11.94\%    \\
                            & AddSent                 &  & 90.92\%       & 100.00\%       &  & 90.99\%     & 12.88\%    \\
                            & Hidden Killer           &  & 90.24\%       & 89.76\%        &  & 90.10\%     & 35.66\%    \\ \midrule
\multirow{3}{*}{DistilBERT} & BadNets                 &  & 88.76\%       & 100.00\%       &  & 90.25\%     & 8.78\%     \\
                            & AddSent                 &  & 89.11\%       & 100.00\%       &  & 89.22\%     & 8.33\%     \\
                            & Hidden Killer           &  & 88.28\%       & 90.18\%        &  & 88.90\%     & 31.28\%    \\ \midrule
\multirow{3}{*}{RoBERTa}    & BadNets                 &  & 93.23\%       & 100.00\%       &  & 93.21\%     & 5.63\%     \\
                            & AddSent                 &  & 92.32\%       & 100.00\%       &  & 92.66\%     & 5.86\%     \\
                            & Hidden Killer           &  & 92.08\%       & 92.74\%        &  & 91.77\%     & 29.07\%    \\ \midrule
\multirow{3}{*}{ALBERT}     & BadNets                 &  & 91.94\%       & 99.77\%        &  & 91.74\%     & 7.65\%     \\
                            & AddSent                 &  & 91.63\%       & 100.00\%       &  & 90.59\%     & 6.08\%     \\
                            & Hidden Killer           &  & 90.96\%       & 88.35\%        &  & 90.27\%     & 38.96\%    \\ \bottomrule
\end{tabular}}
\end{minipage}
\end{table*}

\smallskip
\noindent{\bf{Detailed Detected Triggers.}} 
To further understand the performance of \sys, in \autoref{tab:pr}, we also demonstrate the injected trigger (ground-truth) and the detected trigger of our method. 
The injected triggers for word level attack (BadNets~\cite{gu2017badnets}), sentence level attack (AddSent~\cite{dai2019backdoor}), syntactic level attack (Hidden Killer~\cite{qi2021hidden}), and style level attack (Qi et al.~\cite{qi2021mind}) are word ``tq", sentence ``I watch this 3D movie" , syntactic structure ``S -\textgreater SBAR \_ NP VP \_", and style ``Bible-style", respectively.
In all cases, the injected triggers are successfully found by our method. On the other hand, our method does not have false positive (detecting benign elements as backdoor triggers) in 11/12 settings.
For word level detection on HSOL dataset~\cite{davidson2017automated}, three non-injected words (``bi*ch'',``h*e'' and ``pu*sy'') are also detected as backdoor triggers. This is understandable because these words have high duplication frequency and strong correlation to the label ``offensive".
Existing work~\cite{wallace2019universal,liu2019abs,tao2022backdoor} found that the model trained on benign dataset can also have backdoors. Such backdoors are caused by the bias of the dataset and is called as ``natural backdoor".
Words ``bi*ch'',``h*e'' and ``pu*sy'' here are actually ``natural backdoor" defined in \citet{tao2022backdoor}. Overall, our method achieves better performance than existing NLP backdoor defenses. It can also identify the ``natural backdoor" which is related to the bias of NLP models. 

\smallskip
\noindent{\bf{Precision and Recall of Backdoor Trigger Detection.}} 
We then study the detailed precision and recall of our backdoor trigger detection method. The datasets used here are SST-2~\cite{socher2013recursive}, HSOL~\cite{davidson2017automated} and AG's News~\cite{zhang2015character}. Four attacks (i.e., BadNets~\cite{gu2017badnets}, AddSent~\cite{dai2019backdoor}, Hidden Killer~\cite{qi2021hidden}, and Style~\cite{qi2021mind}) are included. The model used is BERT. In detail, for each attack on each dataset, we run our method 10 times with different random seeds, and get the detailed trigger detection precisions and recalls for the 10 runs. The results can be found in \autoref{tab:detailed_pr}. As can be observed, our method achieves 100.0\% recalls in all cases, meaning that it is highly effective for detecting backdoor triggers. It only has relatively low precisions on BadNets attack and HSOL dataset. As we explained in \autoref{sec:effe}, this is because there are some strongly biased words (e.g., “bi*ch”, “h*e” and “pu*sy”) in this dataset and these strongly biased words are detected as backdoor triggers by our method. As shown in \autoref{tab:compare}, removing such biased words will not have significant influences on the BA and ASR of our method. As long as the recall value for the backdoor detection process is high, our method will have state-of-the-art defense performance even though the detection accuracy is relatively lower.

\begin{table}[]
\scriptsize
\caption{Detailed precisions and recalls of backdoor trigger detection.}
\label{tab:detailed_pr}
\centering
\setlength\tabcolsep{3pt}
\scalebox{1}{
\begin{tabular}{@{}cccc@{}}
\toprule
Dataset                    & Attack        & Precision & Recall  \\ \midrule
\multirow{4}{*}{SST-2}     & BadNets       & 100.0\%   & 100.0\% \\
                           & AddSent       & 100.0\%   & 100.0\% \\
                           & Hidden Killer & 100.0\%   & 100.0\% \\
                           & Style         & 100.0\%   & 100.0\% \\ \midrule
\multirow{4}{*}{HSOL}      & BadNets       & 26.3\%    & 100.0\% \\
                           & AddSent       & 100.0\%   & 100.0\% \\
                           & Hidden Killer & 100.0\%   & 100.0\% \\
                           & Style         & 100.0\%   & 100.0\% \\ \midrule
\multirow{4}{*}{AG’s News} & BadNets       & 100.0\%   & 100.0\% \\
                           & AddSent       & 100.0\%   & 100.0\% \\
                           & Hidden Killer & 100.0\%   & 100.0\% \\
                           & Style         & 100.0\%   & 100.0\% \\ \bottomrule
\end{tabular}}
\end{table}

\section{Robustness to Different Triggers}
\label{sec:appendix_trigger}
To study \sys's robustness to different trigger patterns, 
we show the performance of \sys under different backdoor triggers.
The trigger used here are ``tq", ``cf", ``mn" and ``bb".
The results under different triggers are shown in \autoref{tab:robust_trigger}. For all triggers, our method have low ASR with the BA nearly unchanged compared to undefended models, showing that out method have good robustness to different trigger patterns.

\begin{table}[H]
\centering
\scriptsize
\setlength\tabcolsep{4pt}
\caption{Robustness on different trigger pattern.}\label{tab:robust_trigger}
\begin{tabular}{@{}ccccccc@{}}
\toprule
\multirow{2}{*}{Trigger} &  & \multicolumn{2}{c}{Undefended} &  & \multicolumn{2}{c}{BMC} \\ \cmidrule(lr){3-4} \cmidrule(l){6-7} 
                         &  & BA            & ASR            &  & BA         & ASR        \\ \midrule
tq                       &  & 91.15\%       & 100.00\%       &  & 90.72\%    & 11.94\%    \\
cf                       &  & 91.06\%       & 100.00\%       &  & 90.63\%    & 10.75\%    \\
mn                       &  & 91.28\%       & 100.00\%       &  & 90.88\%    & 11.17\%    \\
bb                       &  & 91.17\%       & 100.00\%       &  & 90.79\%    & 10.86\%    \\ \bottomrule
\end{tabular}
\end{table}

\section{Comparison to More Methods}
\label{sec:appendix_moderate}

In this section, we conduct experiments to compare our method to more training-time backdoor defense methods.
We first compare our method to Moderate-fitting~\cite{zhu2022moderate}. The dataset and the model used are SST-2~\cite{socher2013recursive} and RoBERTa~\cite{liu2019roberta}, respectively. The results can be found in \autoref{tab:compare_zhu}.
The attacks used are BadNets~\cite{gu2017badnets} and Hidden Killer~\cite{qi2021hidden}. As can be seen, our method achieves lower attack success rate (ASR) and higher benign accuracy (BA), meaning that our method outperforms existing method Moderate-fitting~\cite{zhu2022moderate}.
There are also other methods. BFClass~\cite{li2021bfclass} employs a pre-trained discriminator to detect potential replacement tokens, treating them as possible triggers. Similar to BKI~\cite{chen2021mitigating}, it is constrained to word triggers, whereas our methodology is adaptable to various trigger types.
WeDef~\cite{jin2022wedef} trains a weakly-supervised model to identify backdoor samples based on the alignment of weak classifier predictions with their training dataset labels. Its fundamental assumption is the incorrect labeling of poisoned samples, rendering it ineffective against clean-label attacks~\cite{cui2022unified}—a critical category of backdoor assaults. Our approach, in contrast, is demonstrated to be effective against clean-label attacks, as detailed in \autoref{tab:robust_label}.

\begin{table}[H]
\centering
\scriptsize
\setlength\tabcolsep{4pt}
\caption{Comparison to Moderate-fitting~\cite{zhu2022moderate}.}\label{tab:compare_zhu}
\begin{tabular}{@{}cccc@{}}
\toprule
Attack                         & Method           & BA      & ASR     \\ \midrule
\multirow{2}{*}{BadNets}       & Moderate-fitting & 92.71\% & 11.08\% \\
                               & BMC (Ours)       & 93.21\% & 5.63\%  \\ \midrule
\multirow{2}{*}{Hidden killer} & Moderate-fitting & 91.64\% & 41.85\% \\
                               & BMC (Ours)       & 91.77\% & 29.07\% \\ \bottomrule
\end{tabular}
\end{table}

\section{Robustness to Different Label Settings in Poisoning}
\label{sec:appendix_label}

In backdoor attacks, there are different label settings (i.e., dirty-label, clean-label and mix-label) when poisoning the datasets. For dirty-label setting~\cite{liu2017neural}, the labels of the backdoor samples are modified to the target labels, meaning that modified labels are different to the original labels of the samples. For clean-label setting~\cite{turner2019label}, the attackers poison the dataset via pasting the triggers on the samples belong to the target label. In this case, the label of the poisoning samples is identical to their original labels, which increasing the stealthiness of the poisoning. Mix-label setting means half of the poisoning samples are in dirty-label setting and another half is in clean-label setting.
To investigate \sys's robustness on different label settings in poisoning, we show the BA and ASR of the models
 under different label settings. 
 As can be observed, our method achieves good defense performances (i.e., low ASR and high BA) under all different label settings, showing the robustness to different label settings in poisoning.

 \begin{table}[]
\centering
\scriptsize
\setlength\tabcolsep{4pt}
\caption{Robustness on different label setting.}\label{tab:robust_label}
\begin{tabular}{@{}ccccccc@{}}
\toprule
\multirow{2}{*}{Label Setting} &  & \multicolumn{2}{c}{Undefended} &  & \multicolumn{2}{c}{BMC} \\ \cmidrule(lr){3-4} \cmidrule(l){6-7} 
                               &  & BA            & ASR            &  & BA         & ASR        \\ \midrule
Dirty-label                    &  & 91.62\%       & 100.00\%       &  & 90.93\%    & 9.46\%     \\
Mixed-label                    &  & 91.15\%       & 100.00\%       &  & 90.72\%    & 11.94\%    \\
Clean-label                    &  & 91.28\%       & 96.39\%        &  & 90.19\%    & 10.82\%    \\ \bottomrule
\end{tabular}
\end{table}

\section{Efficiency}
\label{sec:efficiency}

We propose a variety of accelerated search methods to search potential backdoor samples efficiently, which helps the application of the methods in practice (see \autoref{sec:method}). These accelerated search methods have good theoretical computing complexity and scalable to larger magnitudes of training data as they are based on efficient data structures and algorithms such as Hash Table and Suffix Array. 
To measure the efficiency of different method, we report the runtime for different training-time NLP backdoor defenses, i.e., BKI, CUBE, and BMC. The results are shown in \autoref{tab:time_compare}. 
The results demonstrate that the runtime of our method is comparable to that of existing methods. Currently, the main runtime bottleneck is brought from the syntax structure trigger verification stage, which is based on the existing structure transfer method Iyyer et al.~\cite{iyyer2018adversarial} . It can be solved by proposing more efficient structure transfer methods or using parallel computing techniques, which are orthogonal to the goal of this paper.

\begin{table}[]
\centering
\scriptsize
\setlength\tabcolsep{4pt}
\caption{Runtime for different method.}\label{tab:time_compare}
\begin{tabular}{@{}cc@{}}
\toprule
Method               & Runtime \\ \midrule
Standard Fine-tuning & 126.66s \\
BKI                  & 499.53s \\
CUBE                 & 433.74s \\
BMC                  & 473.85s \\ \bottomrule
\end{tabular}
\end{table}

\section{Discussion about Trigger Positions}
\label{sec:position}

 Based on our knowledge, existing NLP backdoor attacks are position-agnostic. More specifically, when the attackers inject such a backdoor, the triggers' positions can be random, and the attack will still be successful. We acknowledge that there are efforts to make such attacks position-specific. Li et al.~\cite{li2021hiddennlp} studied the NLP backdoor attack’s transferability of trigger positions, demonstrating that the learned position conditions are not yet accurate. Namely, the backdoored model poisoned on one specific trigger position (e.g., rear of the sample) can be effectively activated by a different position trigger (e.g., front of the sample). Due to many reasons (e.g., sparsity of NLP input domain, various lengths of NLP inputs), position-specific backdoor attacks in data poisoning scenario remain challenging. Thus, following existing work, in our default setting, we insert the candidate triggers at random positions in the original sample during the trigger verification stage.
 Developing position-specific backdoor attacks and corresponding defenses will be our future work.
\section{Adaptive Attack}
\label{sec:adaptive}
In this section, we discuss the potential adaptive attacks.
To the best of our knowledge, all successful data-poisoning based NLP backdoor attacks will reflect their triggers on the instantialized syntax tree, and our approach is general for the attacks in our threat model. It is hard to do an adaptive attack without full control of the training process. Since our method is based on the duplication frequency of the trigger elements, a method with extremely low poisoning rates can be viewed as the adaptive attack for our method. We have demonstrated the results under extremely low poisoning rates (e.g., 0.85\%) in \autoref{tab:robust_pr}, and the results show that our method is robust to this adaptive attack. Note that the attacks with poisoning rates lower than 0.85\% will yield low attack success rates (See \autoref{fig:pr}), and 0.85\% is actually extremely low for successful attacks.

\section{More Details about the Related Backdoor Attacks \& Defenses}
\label{sec:detailed_related}
\para{Backdoor Attacks} 
Machine learning models are vulnerable to backdoors~\cite{gu2017badnets,liu2017trojaning,turner2019label}.
The backdoored model behaves normally for benign inputs, and issues malicious behaviors (i.e., predicting a certain target label) when the input is stamped with the backdoor trigger (i.e., a specific input pattern).
One common method to inject backdoors is training data poisoning~\cite{gu2017badnets,turner2019label,chen2021badnl}, where attackers add triggers to training samples to create a strong connection between the trigger and a target label of their choosing. In the field of natural language processing (NLP), existing works~\cite{gu2017badnets,chen2021badnl,kurita2020weight,dai2019backdoor} have proposed different methods for inserting triggers into training data, such as using unusual words~\cite{gu2017badnets,chen2021badnl,kurita2020weight} or sentences~\cite{dai2019backdoor}.  
Prior works also proposed more stealthy triggers, such as modifying syntactic structure~\cite{qi2021hidden} and transferring text style~\cite{qi2021mind, pan2022hidden}, which are less noticeable.

\para{Backdoor Defenses} 
The growing concern of backdoor attacks has led to the development of various defense approaches to prevent them~\cite{wang2019neural,tran2018spectral,pmlr-v139-hayase21a,gao2019strip,qi2021onion,wang2022rethinking,liu2022piccolo,jin2023incompatibility,yang2024robust}. In NLP, the defense methods for backdoor attacks can be broadly grouped into three categories.
Training-time defenses~\cite{chen2021mitigating,cui2022unified} include methods like CUBE~\cite{cui2022unified}, which eliminate suspicious samples with potential backdoor keywords from the training data. 
Both CUBE~\cite{cui2022unified} and our method are categorized as training-time defenses, aiming to train clean models from a poisoned training dataset. The principal distinction of our method from existing training-time defense strategies like CUBE is our focus on the properties of training data and their interplay with language model memorization, rather than analyzing internal behaviors of the language model (such as weights and activation values). Rooted in our theoretical exploration of the connection between backdoor behaviors and language model memorization, our approach demonstrates enhanced efficacy and robustness compared to other training-time defenses that primarily rely on heuristic observations.
Runtime defenses~\cite{gao2021design,yang2021rap} include techniques such as RAP~\cite{yang2021rap} and STRIP~\cite{gao2019strip}, which remove words that may be backdoor triggers from testing samples and using robustness-aware perturbations to distinguish poisoned data from clean data during the inference stage. 
 These methods differ significantly in their threat models compared to our approach. Their objective is to distinguish between backdoor and clean samples during the model's inference phase, unlike our method, which is focused on the training stage.
Another way to defend against the NLP backdoor is to detect if a given model is infected with the backdoor or not via backdoor trigger inversion~\cite{liu2022piccolo,shen2022constrained,azizi2021t}. This kind of defense requires a set of clean samples to conduct trigger reverse engineering, limiting its practicality.

\begin{table}[]
\centering
\scriptsize
\setlength\tabcolsep{4pt}
\caption{Results on generative model and question answering task.}\label{tab:generative}
\begin{tabular}{@{}ccc@{}}
\toprule
Method     & Quality & ASR    \\ \midrule
Undefended & 5.0     & 44.5\% \\
Ours       & 4.9     & 0.0\%  \\ \bottomrule
\end{tabular}
\end{table}

\section{Extension to Generative Tasks and Generative Models}
\label{sec:extension_tasks}

In this section, we discuss our method's application on the generative tasks and the generative models. The attack used here is 
\citet{yan2023virtual}.
While most of the existing data-poisoning-based NLP backdoor attacks focus on the classification tasks~\cite{chen2021badnl,kurita2020weight,dai2019backdoor,qi2021mind, pan2022hidden}, there are also an attack focusing on the generative models such as Alpaca~\cite{taori2023stanford}.
Therefore, we conduct the experiments on this attack in this section. The model used here is Alpaca 7B~\cite{taori2023stanford} (a instruction tuned large language model based on LLaMA~\cite{touvron2023llama}). The instruction dataset is generated using the procedure described in \citet{yan2023virtual} with trigger ``Joe Biden''. The target of the attack here is the negative sentiment presented in the generated answers. The poisoning rate here is 1\%. The metrics used is the attack success rate (ASR) and the generation quality measured by GPT-4 rating on a scale of 1 to 10~\cite{yan2023virtual}.
The results are shown in \autoref{tab:generative}. As can be observed, our method can effectively reduce the ASR while keep the generation quality of the protected model.
Thus, our method also have satisfying performance on the generative tasks such as question answering and the generative models such as Alpaca.

\end{document}